\documentclass{article}

\usepackage{microtype}

\usepackage{graphicx}
\usepackage{tikz}
\usetikzlibrary{positioning}
\usepackage{subfigure}
\usepackage{booktabs}
\usepackage{todonotes}

\usepackage{hyperref}

\usepackage{amsthm, amssymb, amsmath}
\usepackage{mathtools}                  
\usepackage{dsfont}

\newtheorem{theorem}{Theorem}[section]

\theoremstyle{definition}
\newtheorem{definition}{Definition}

\newtheorem{remark}[definition]{Remark}

\DeclareMathOperator*{\E}{\mathbb{E}}
\DeclareMathOperator*{\R}{\mathbb{R}}
\DeclareMathOperator*{\argmin}{arg\,min}

\usepackage{siunitx}
\newcommand{\review}[1]{\textcolor{black}{#1}}

\usepackage[accepted]{icml2021}

\icmltitlerunning{Solving high-dimensional parabolic PDEs using the tensor train format}

\begin{document}

\twocolumn[
\icmltitle{Solving high-dimensional parabolic PDEs using the tensor train format}

\icmlsetsymbol{equal}{*}

\begin{icmlauthorlist}
\icmlauthor{Lorenz Richter}{equal,FU,BTU,dida}
\icmlauthor{Leon Sallandt}{equal,TU}
\icmlauthor{Nikolas N\"usken}{potsdam}
\end{icmlauthorlist}

\icmlaffiliation{FU}{Freie Universit\"at Berlin, Germany}
\icmlaffiliation{BTU}{BTU Cottbus-Senftenberg, Germany}
\icmlaffiliation{dida}{dida Datenschmiede GmbH, Germany}
\icmlaffiliation{TU}{Technische Universit\"at Berlin, Germany}
\icmlaffiliation{potsdam}{Universit\"at Potsdam, Germany}

\icmlcorrespondingauthor{Lorenz Richter}{lorenz.richter@fu-berlin.de}
\icmlcorrespondingauthor{Leon Sallandt}{sallandt@math.tu-berlin.de}

\icmlkeywords{Machine Learning, ICML}

\vskip 0.3in
]

\printAffiliationsAndNotice{\icmlEqualContribution} 

\begin{abstract}
High-dimensional partial differential equations (PDEs) are ubiquitous in economics, science and engineering. However, their numerical treatment poses formidable challenges since traditional grid-based methods tend to be frustrated by the curse of dimensionality. In this paper, we argue that tensor trains provide an appealing approximation framework for parabolic PDEs: the combination of reformulations in terms of backward stochastic differential equations and regression-type methods in the tensor format holds the promise of leveraging latent low-rank structures enabling both compression and efficient computation. Following this paradigm, we develop novel iterative schemes, involving either explicit and fast or implicit and accurate updates. We demonstrate in a number of examples that our methods  achieve a favorable trade-off between accuracy and computational efficiency in comparison with state-of-the-art neural network based approaches.

\end{abstract}

\section{Introduction}
\label{sec: introduction}
While partial differential equations (PDEs) offer one of the most elegant frameworks for modeling in economics, science and engineering, their practical use is often limited by the fact that solving those equations numerically becomes notoriously difficult in high-dimensional settings. The so-called ``curse of dimensionality'' refers to the phenomenon that the computational effort scales exponentially in the dimension, rendering classical grid based methods infeasible. In recent years there have been fruitful developments in combining Monte Carlo based algorithms with neural networks in order to tackle high-dimensional problems in a way that seemingly does not suffer from this curse, resting primarily on stochastic representations of the PDEs under consideration \cite{weinan2017deep, raissi2019physics, weinan2019multilevel, hure2020deep, nusken2020solving}. Many of the suggested algorithms perform remarkably well in practice and some theoretical results proving beneficial approximation properties of neural networks in the PDE setting are now available \cite{jentzen2018proof}. Still, a complete picture remains elusive, and the optimization aspect in particular continues to pose challenging and mostly open problems, both in terms of efficient implementations and theoretical understanding. Most importantly for practical applications, neural network training using gradient descent type schemes may often take a very long time to converge for complicated PDE problems.

Instead of neural networks (NN), we propose relying on the tensor train (TT) format 
\cite{oseledets2011tensor} to approximate the solutions of high-dimensional PDEs.
As we argue in the course of this article, the salient features of tensor trains make them an ideal match for the stochastic methods alluded to in the previous paragraph: 
First, tensor trains have been designed to tackle high-dimensional problems while still being computationally cheap by exploiting inherent low-rank structures \cite{kazeev2012low, kazeev2016qtt, dolgov2012fast} typically encountered in  physically inspired PDE models. Second, built-in orthogonality relations allow fast and robust optimization in regression type problems arising naturally in stochastic backward formulations of parabolic PDEs. Third, the function spaces corresponding to tensor trains can be conveniently extended to incorporate additional information such as initial or final conditions imposed on the PDE to be solved. Last but not least, tensor trains allow for extremely efficient and explicit computation of first and higher order derivatives. 

To develop TT-based solvers for parabolic PDEs, we follow \cite{bouchard2004discrete, hure2020deep} and first identify a backward stochastic differential equation (BSDE) representation of the PDE, naturally giving rise to iterative backward schemes for a numerical treatment. We suggest two versions of our algorithm, allowing to adjust the trade-off between accuracy and speed according to the application:
The first scheme is explicit, relying on $L^2$ projections \cite{gobet2005regression} that can be solved efficiently using an alternating least squares algorithm and explicit expressions for the minimizing parameters (see Section \ref{sec: optimization on TT manifold}). The second scheme is implicit and involves a nested iterative procedure, holding the promise of more accurately resolving highly nonlinear relationships at the cost of an increased computational load. For theoretical underpinning, we prove the convergence of the nested iterative scheme in Section \ref{sec: handling non-linear equations}. 

To showcase the performance of the TT-schemes, we evaluate their outputs on various high-dimensional PDEs (including toy examples and real-world problems) in comparison with  NN-based approximations.  In all our examples, the TT results prove competitive, and often considerably more accurate when low-rank structures can be identified and captured by the underlying ansatz spaces.  At the same time, the runtimes of the TT-schemes are usually significantly smaller, with the explicit $L^2$-projection-based algorithm beating the corresponding NN alternative by orders of magnitude in terms of computational time. Even the more accurate algorithm based on nested nonlinear iterations often proves to be substantially faster than NN training.

\subsection{Previous work}
Using numerical discretizations of BSDEs to solve PDEs originated in \cite{bouchard2004discrete, gobet2005regression}, while regression based methods for PDE-related problems in mathematical finance have already been proposed in \cite{longstaff2001valuing}. An iterative method motivated by BSDEs and approached with neural networks has been introduced in \cite{weinan2017deep}, making the approximation of high-dimensional PDE problems feasible. Solving explicit backwards schemes with neural networks has been suggested in \cite{beck2019deep} and an implicit method similar to the one developed in this paper has been suggested in \cite{hure2020deep}. Another interesting method to approximate PDE solutions relies on minimizing a residual term on uniformly sampled data points as suggested in \cite{sirignano2018dgm, raissi2019physics}.
Rooted in quantum physics under the name \emph{matrix product states}, tensor trains have been introduced to the mathematical community in \cite{oseledets2011tensor} to tackle the curse of dimensionality.
\review{Note that tensor trains are a special case of hierarchical tensor networks, which have been developed in \cite{Hackbusch-2010}.}
For good surveys and more details, see \cite{Hackbusch-Acta, Hackbusch2014,Legeza-Schneider,Bachmayr-Uschmajew-Schneider}.
\review{Tensor trains have already been applied to parametric PDEs, see e.g. \cite{dolgov2015polynomial, eigel2017adaptive, dektor2020rank}, Hamilton-Jacobi-Bellman PDEs \cite{horowitz2014linear, stefansson2016sequential, gorodetsky2018high,dolgov2019tensor,oster2019approximating,fackeldey2020approximative, chen2021tensor},
and PDEs of other types, see e.g. \cite{khoromskij2012tensors, kormann2015semi, lubasch2018multigrid}.
}

The paper is organized as follows: In Section \ref{sec: solving PDEs via BSDEs} we motivate our algorithm by recalling the stochastic PDE representation in terms of BSDEs as well as two appropriate discretization schemes. In Section \ref{sec: tensor trains} we review the tensor train format as a highly efficient framework for approximating high-dimensional functions by detecting low-rank structures and discuss how those structures can be exploited in the numerical solution of BSDEs. Finally, in Section \ref{sec: numerical examples} we provide multiple high-dimensional numerical examples to illustrate our claims.

\section{Solving PDEs via BSDEs}
\label{sec: solving PDEs via BSDEs}

In this section we recall how backward stochastic differential equations (BSDEs) can be used to design iterative algorithms for approximating the solutions of high-dimensional PDEs. Throughout this work, we consider parabolic PDEs of the form
\begin{equation}
\label{eq: definition general PDE}
    (\partial_t + L) V(x, t) + h(x, t, V(x, t), (\sigma^\top \nabla V)(x, t)) = 0
\end{equation}
for $(x, t) \in {\R}^d \times [0, T]$, a nonlinearity $h: \R^d \times [0, T] \times \R \times \R^d \to \R $, and a differential operator 
\begin{equation}
\label{eq: infinitesimal generator}
    L = \frac{1}{2} \sum_{i,j=1}^d (\sigma \sigma^\top)_{ij}(x,t) \partial_{x_i} \partial_{x_j} + \sum_{i=1}^d b_i(x,t) \partial_{x_i},
\end{equation}
with coefficient functions $b:\R^d \times [0, T] \to \R^d$ and $\sigma:\R^d \times [0, T] \to \R^{d \times d}$. The terminal value is given by
\begin{equation}
\label{eq: definition boundary value}
    V(x, T) = g(x),
\end{equation}
for a specified function $g : \R^d \to \R$. Note that by using the time inversion $t \mapsto T - t$, the terminal value problem \eqref{eq: definition general PDE}-\eqref{eq: definition boundary value} can readily be transformed
into an initial value problem.

BSDEs were first introduced in \cite{bismut1973conjugate} and their systematic study began with \cite{pardoux1990adapted}. Loosely speaking, they can be understood as  nonlinear extensions of the celebrated Feynman-Kac formula \cite{pardoux1998backward}, relating the PDE \eqref{eq: definition general PDE} to the stochastic process $X_s$ defined by
\begin{equation}
\label{eq: fordward SDE}
    \mathrm dX_s = b(X_s, s) \, \mathrm ds + \sigma(X_s, s) \, \mathrm d W_s, \quad X_0 = x_0,
\end{equation}
where $b$ and $\sigma$ are as in \eqref{eq: infinitesimal generator} and $W_s$ is a standard $d$-dimensional Brownian motion. The key idea is then to define the processes
\begin{equation}
\label{eq: def Y Z}
    Y_s = V(X_s, s), \qquad Z_s = (\sigma^\top \nabla V)(X_s, s)
\end{equation}
as representations of the PDE solution and its gradient,
and apply It\^{o}'s lemma to obtain
\begin{equation}
\label{eq: BSDE}
    \mathrm d Y_s  = -h(X_s, s, Y_s, Z_s) \, \mathrm ds +  Z_s \cdot \mathrm dW_s, 
\end{equation}
with terminal condition $Y_T  = g(X_T)$.
Noting that the processes $Y_s$ and $Z_s$ are adapted\footnote{Intuitively, this means that the processes $Y_s$ and $Z_s$ must not depend on future values of the Brownian motion $W_s$.} to the filtration generated by the Brownian motion $W_s$, they should indeed be understood as backward processes and not be confused with time-reversed processes. A convenient interpretation of the relations in \eqref{eq: def Y Z} is that solving for the processes $Y_s$ and $Z_s$
under the constraint \eqref{eq: BSDE} corresponds to determining the solution of the PDE \eqref{eq: definition general PDE} (and its gradient) along a random grid which is provided by the stochastic process $X_s$ defined in \eqref{eq: fordward SDE}.

\subsection{Numerical approximation of BSDEs}
\label{sec: numerical approximation of BSDEs}
The BSDE formulation \eqref{eq: BSDE} opens the door for Monte Carlo algorithms aiming to numerically approximate $Y_s$ and $Z_s$, and hence yielding approximations of solutions to the PDE \eqref{eq: definition general PDE} according to \eqref{eq: def Y Z}, see \cite{bouchard2004discrete, gobet2005regression}. In this section we discuss suitable discretizations of \eqref{eq: BSDE} and corresponding optimization problems that will provide the backbone for TT-schemes to be developed in Section \ref{sec: tensor trains}.   

To this end, let us define a discrete version of the process \eqref{eq: fordward SDE} on a time grid $0 = t_0 < t_1 < \dots < t_N = T$ by
\begin{equation}
\label{eq: discrete SDE}
 \widehat{X}_{n+1} = \widehat{X}_n + b(\widehat{X}_n, t_n) \Delta t + \sigma(\widehat{X}_n, t_n) \xi_{n+1} \sqrt{\Delta t},
\end{equation}
where $n \in \{0, \dots, N-1\}$ enumerates the steps, $\Delta t = t_{n+1} - t_n$ is the stepsize, $\xi_{n+1} \sim \mathcal{N}(0, \text{Id}_{d\times d})$ are normally distributed random variables and $\widehat{X}_0 = x_0$ provides the initial condition. Two\footnote{It can be shown that both converge to the continuous-time process \eqref{eq: BSDE} as $\Delta t \rightarrow 0$, see \cite{kloeden1992stochastic}.} discrete versions of the backward process \eqref{eq: BSDE} are given by
\begin{subequations}
\label{eq: backward schemes}
\begin{align}
    \label{eq: discrete BSDE explicit}
    \widehat{Y}_{n+1} &= \widehat{Y}_n - h_{n+1} \Delta t +  \widehat{Z}_n \cdot \xi_{n+1}  \sqrt{\Delta t},
\\
    \label{eq: discrete BSDE}
    \widehat{Y}_{n+1} &= \widehat{Y}_n - h_n \Delta t +  \widehat{Z}_n \cdot \xi_{n+1}  \sqrt{\Delta t},
\end{align}
\end{subequations}
where we have introduced the shorthands
\begin{subequations}
\begin{align}
\label{eq: shorthand notation h_n}
    h_n &= h(\widehat{X}_n, t_n, \widehat{Y}_n, \widehat{Z}_n), \\
    h_{n+1} &= h(\widehat{X}_{n+1}, t_{n+1}, \widehat{Y}_{n+1}, \widehat{Z}_{n+1}).
    \label{eq: shorthand notation h_n_1}
\end{align}
\end{subequations}
Finally, we complement \eqref{eq: discrete BSDE explicit} and \eqref{eq: discrete BSDE}  by specifying the terminal condition $\widehat{Y}_N = g(\widehat{X}_N)$. The reader is referred to Appendix \ref{app: BSDE background} for further details.

Both of our schemes solve the discrete  processes \eqref{eq: discrete BSDE explicit} and \eqref{eq: discrete BSDE} backwards in time, an approach which is reminiscent of the dynamic programming principle in optimal control theory \cite{fleming2012deterministic}, where the problem is divided into a sequence of subproblems. To wit, we start with the known terminal value $\widehat{Y}_N = g(\widehat{X}_N)$ and move backwards in iterative fashion until reaching $\widehat{Y}_0$. Throughout this procedure, we posit functional approximations $\widehat{V}_n(\widehat{X}_n) \approx \widehat{Y}_n \approx V(\widehat{X}_n,t_n)$ to be learnt in the update step $n + 1 \rightarrow n$  which can either be based on \eqref{eq: discrete BSDE explicit} or on \eqref{eq: discrete BSDE}:

Starting with the former, it can be shown by leveraging the relationship between conditional expectations and $L^2$-projections (see Appendix \ref{app: BSDE background}) that solving \eqref{eq: discrete BSDE explicit} is equivalent to minimizing  
\begin{equation}
\label{eq: projection_based_optimization}
\E\left[ \left(\widehat{V}_n(\widehat{X}_n) -h_{n+1} \Delta t - \widehat{V}_{n+1}(\widehat{X}_{n+1})\right)^2\right]
\end{equation}
with respect to  $\widehat{V}_n$. Keeping in mind that $\widehat{V}_{n+1}$ is known from the previous step this results in an explicit scheme. Methods based on \eqref{eq: projection_based_optimization} have been extensively analyzed in the context of linear ansatz spaces for $\widehat{V}_n$ and we refer to \cite{zhang2004numerical, gobet2005regression} as well as to Appendix \ref{app: BSDE background}.

Moving on to \eqref{eq: discrete BSDE}, we may as well penalize deviations in this relation by minimizing the alternative loss
\begin{multline}
\label{eq: implicit scheme}
    \E[ (\widehat{V}_n(\widehat{X}_n) - \widehat{h}_n \Delta t - \widehat{V}_{n+1}(\widehat{X}_{n+1}) \\
    + \sigma^\top(\widehat{X}_n, t_n) \nabla \widehat{V}_n(\widehat{X}_n) \cdot \xi_{n+1}\sqrt{\Delta t})^2  ],
\end{multline}

with respect to $\widehat{V}_n$, see \cite{hure2020deep}. In analogy to \eqref{eq: shorthand notation h_n} we use the shorthand notation
\begin{equation}
    \widehat{h}_n = h(\widehat{X}_n, t_n, \widehat{V}_n(\widehat{X}_n) , \sigma^\top(\widehat{X}_n, t_n) \nabla \widehat{V}_n(\widehat{X}_n) ), 
\end{equation}
noting that since $\widehat{h}_n$
depends on $\widehat{V}_n$, approaches based on \eqref{eq: implicit scheme} will necessarily lead to implicit schemes. At the same time, we expect algorithms based on \eqref{eq: implicit scheme} to be more accurate in highly nonlinear scenarios as the dependence in $h$ is resolved to higher order.

\section{Solving BSDEs via tensor trains}
\label{sec: tensor trains}

In this section we discuss the functional approximations $\widehat{V}_n$ in terms of the tensor train format, leading to efficient optimization procedures for \eqref{eq: projection_based_optimization} and \eqref{eq: implicit scheme}.
Encoding  functions defined on high-dimensional spaces using traditional methods such as finite elements, splines or multi-variate polynomials leads to a computational complexity that scales exponentially in the state space dimension $d$.
However, interpreting the coefficients of such ansatz functions as entries in a high-dimensional tensor allows us to use tensor compression methods to reduce the number of parameters.
To this end, we define a set of functions $\{ \phi_1, \dots, \phi_m \}$ with $\phi_i : \mathbb R \to \mathbb R$ ,
e.g. one-dimensional polynomials or finite elements.
The approximation $\widehat V$ of $V:\mathbb{R}^d \rightarrow \mathbb{R}$ takes the form
\begin{equation}
\label{eq:V c}
    \widehat{V}(x_1, \dots, x_d) = \sum_{i_1 = 1}^m \dots \sum_{i_d = 1}^m c_{i_1, \dots, i_d} \phi_{i_1}(x_1) \cdots \phi_{i_d} (x_d),
\end{equation}
motivated by the fact that polynomials and other tensor product bases are dense in many standard function spaces \cite{sickel2009tensor}.
Note that for the sake of simplicity we choose the set of ansatz functions to be the same in every dimension (see Appendix \ref{sect:graphical_notation} for more general statements). 
As expected, the coefficient tensor $c \in \mathbb R^{m \times m \times \dots \times m} \equiv \mathbb R^{m^d}$ suffers from the curse of dimensionality since the number of entries increases exponentially in the dimension $d$.
In what follows, we review the tensor train format to compress the tensor $c$.

For the sake of readability we will henceforth write $c_{i_1, \dots, i_d} = c[i_1, \dots, i_d]$ and represent the contraction of the last index of a tensor $w_1 \in \mathbb R^{r_1 \times m \times r_2}$ with the first index of another tensor $w_2 \in \mathbb R^{r_2 \times m \times  r_3}$ by
\begin{subequations}
\begin{align}
    w &= w_1 \circ w_2 \in \mathbb R^{r_1 \times m \times m \times r_3}, \\
    w[i_1, i_2, i_3, i_4] &= \sum_{j = 1}^{r_2} w_1[i_1, i_2, j] w_2[j, i_3, i_4].
\end{align}
\end{subequations}
In the literature on tensor methods, graphical representations of general tensor networks are widely used.
In these pictorial descriptions, the contractions $\circ$ of the component tensors are indicated as edges between vertices of a graph.
As an illustration, we provide the graphical representation of an order-$4$ tensor and a tensor train representation (see Definition \ref{def:tensor_train} below) in Figure \ref{TT:fig:hosvd}. Further examples can be found in  Appendix \ref{sect:graphical_notation}.
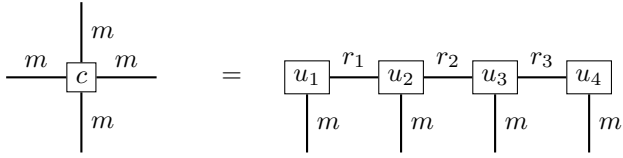
\begin{figure}[h!]
    \centering
    \begin{tikzpicture}
        \begin{scope}[every node/.style={draw,  fill=white}]
        \node (A1) at (0,0) {$u_1$}; 
        \node (A2) at (1.25,0) {$u_2$}; 
        \node (A3) at (2.5,0) {$u_3$}; 
        \node (A4) at (3.75,0) {$u_4$}; 
        
        \node (A0) at (-3,0) {$c$}; 
        \end{scope}
        \node (B2) at (-1,0) {$=$}; 
        \begin{scope}[every edge/.style={draw=black,thick}]
        	\path [-] (A1) edge node[midway,left,sloped] [above] {$r_1$} (A2);
        	\path [-] (A2) edge node[midway,left,sloped] [above] {$r_2$} (A3);
        	\path [-] (A3) edge node[midway,left,sloped] [above] {$r_3$} (A4);
        	
        	\path [-] (A1) edge node[midway,left] [right] {$m$} +(-90:1);
        	\path [-] (A2) edge node[midway,left] [right] {$m$} +(-90:1);
        	\path [-] (A3) edge node[midway,left] [right] {$m$} +(-90:1);
        	\path [-] (A4) edge node[midway,left] [right] {$m$} +(-90:1);
        	
        	\path [-] (A0) edge node[midway,left] [right] {$m$} +(-90:1);
        	\path [-] (A0) edge node[midway,left] [above] {$m$} +(180:1);
        	\path [-] (A0) edge node[midway,left] [right] {$m$} +(90:1);
        	\path [-] (A0) edge node[midway,left] [above] {$m$} +(0:1);
        \end{scope}
    \end{tikzpicture}
    \caption{An order $4$ tensor and a tensor train representation. }
    \label{TT:fig:hosvd}
\end{figure}

Tensor train representations of $c$ can now be defined as follows \cite{oseledets2011tensor}.
\begin{definition}[Tensor Train]\label{def:tensor_train}
    Let $c \in \mathbb R^{m \times \dots \times m}$.
    A factorization
    \begin{equation}
    \label{eq:TT rep}
        c = u_1 \circ u_2 \circ \dots \circ u_d,
    \end{equation}
    where $u_1 \in \mathbb R^{m \times r_1}$, $u_i \in \mathbb R^{r_{i-1} \times m \times r_i}$, $2 \leq i \leq d-1$, $u_d \in \mathbb R^{r_{d-1} \times m}$, is called \emph{tensor train representation} of $c$. 
    We say that $u_i$ are \emph{component tensors}. The tuple of the dimensions $(r_1, \dots, r_{d-1})$ is called the representation rank and is associated with the specific representation \eqref{eq:TT rep}.
    In contrast to that, the tensor train rank (TT-rank) of $c$ is defined as the minimal rank tuple $\mathbf r = (r_1, \dots, r_{d-1})$, such that there exists a TT representation of $c$ with representation rank equal to $\mathbf r$. Here,  minimality of the rank is defined in terms of the partial order relation on $\mathbb N^d$ given by
\[ \mathbf s \preceq \mathbf t \iff s_i \leq t_i \text{ for all } 1 \leq i \leq d,\]
for $\mathbf r = (r_1, \dots, r_d), \, \mathbf s = (s_1, \dots, s_d) \in \mathbb N^d$.
\end{definition}

It can be shown that every tensor has a TT-representation with minimal rank, implying that the TT-rank is well defined \cite{holtz2012manifolds}.
An efficient algorithm for computing a minimal TT-representation is given by the  Tensor-Train-Singular-Value-Decomposition (TT-SVD) \cite{oseledets2009breaking}.
Additionally, the set of tensor trains with fixed TT-rank forms a smooth manifold, and if we include lower ranks, an algebraic variety is formed \cite{landsberg2012tensors, kutschan2018tangent}.

Introducing the compact notation
\[ \phi: \mathbb R \to \mathbb R^{m}, \quad \phi(x) = [\phi_1(x), \dots, \phi_{m}(x) ], \]
the TT-representation of \eqref{eq:V c} is then given as
\begin{multline}
    \widehat V(x) = \sum_{i_1}^m \cdots \sum_{i_d}^m \sum_{j_1}^{r_1} \cdots \sum_{j_{d-1}}^{r_{d-1}} u_1[i_1, j_1] u_2 [j_1, i_2, j_2] \cdots  \\
    \cdots u_d[j_{d-1}, i_d] \phi(x_1)[i_1] \cdots \phi(x_d)[i_d].
\end{multline}
The corresponding graphical TT-representation (with $d=4$ for definiteness) is then given as follows: 
\begin{figure}[h!]
    \centering
    \begin{tikzpicture}
        \begin{scope}[every node/.style={draw,  fill=white}]
        \node (A1) at (0,0) {$u_1$}; 
        \node (A2) at (1.25,0) {$u_2$}; 
        \node (A3) at (2.5,0) {$u_3$}; 
        \node (A4) at (3.75,0) {$u_4$}; 
        
        \node (B1) at (0,-1) {$\phi(x_1)$}; 
        \node (B2) at (1.25,-1) {$\phi(x_2)$}; 
        \node (B3) at (2.5,-1) {$\phi(x_3)$}; 
        \node (B4) at (3.75,-1) {$\phi(x_4)$}; 
        \end{scope}
        \node (C0) at (-2,0) {$\widehat V(x)$}; 
        \node (C1) at (-1,0) {$=$}; 
        \begin{scope}[every edge/.style={draw=black,thick}]
        	\path [-] (A1) edge node[midway,left,sloped] [above] {$r_1$} (A2);
        	\path [-] (A2) edge node[midway,left,sloped] [above] {$r_2$} (A3);
        	\path [-] (A3) edge node[midway,left,sloped] [above] {$r_3$} (A4);
        	\path [-] (A1) edge node[midway,left] [right] {$m$} (B1);
        	\path [-] (A2) edge node[midway,left] [right] {$m$} (B2);
        	\path [-] (A3) edge node[midway,left] [right] {$m$} (B3);
        	\path [-] (A4) edge node[midway,left] [right] {$m$} (B4);
        	
        \end{scope}
    \end{tikzpicture}
    \caption{Graphical representation of $\widehat V: \mathbb R^4 \to \mathbb R$.}
    \label{fig:valuefun1}
\end{figure}
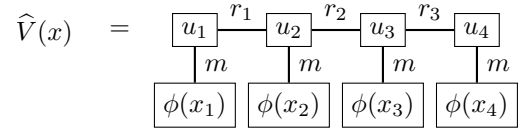

\subsection{Optimization on the TT manifold}
\label{sec: optimization on TT manifold}

The multilinear structure of the tensor product enables efficient optimization of \eqref{eq: projection_based_optimization} and \eqref{eq: implicit scheme} within the manifold structure by means of reducing a high-dimensional linear equation in the coefficient tensor to small linear subproblems on the component tensors\footnote{In the case of \eqref{eq: implicit scheme}, an additional nested iterative procedure is required, see Section \ref{sec: handling non-linear equations}.}. For this, we view \eqref{eq: projection_based_optimization} and \eqref{eq: implicit scheme} abstractly as least squares problems on a linear space $\mathcal{U} \subset L^2(\Omega)$, where $\Omega \subset \mathbb{R}^d$ is a bounded Lipschitz domain. 
Our objective is then to find
\begin{equation}\label{eq:lin_least_squares}
    \argmin_{\widehat V \in \mathcal{U}} \sum_{j=1}^J |\widehat V(x_j) - R(x_j) |^2,
\end{equation}
where $\{x_1, \dots, x_J \} \subset \Omega$ are data points obtained from samples of $\widehat{X}_n$, and $R:\Omega \to \R$ stands for the terms in  \eqref{eq: projection_based_optimization} and \eqref{eq: implicit scheme} that are not varied in the optimization.
Choosing a basis $\{b_1, \dots, b_M \}$ of $\mathcal{U}$ we can represent any function $w \in \mathcal{U}$ by $w(x) = \sum_{m = 1}^M c_m b_m(x)$ and it is well known that the solution to \eqref{eq:lin_least_squares} is given in terms of the coefficient vector
\begin{equation}
\label{eq:regression}
 c = (A^\top A )^{-1}A^\top r \in \mathbb{R}^M,\end{equation}
where $A = [a_{ij}] \in \mathbb R^{J \times M}$ with $a_{ij} = b_j (x_i)$ and $r_j = R(x_j) \in \mathbb R^J$.

The \emph{alternating least-squares (ALS) algorithm}  \cite{ALS} reduces the high-dimensional system \eqref{eq:regression} in the coefficient tensor $c$ to small linear subproblems in the component tensors $u_i$ as follows:
Since the tensor train format \eqref{eq:TT rep} is a multilinear parametrization of $c$, fixing every component tensor but one (say $u_i$) isolates a remaining low-dimensional linear parametrization with associated local linear subspace $\mathcal{U}_{\mathrm{loc},i}$.
The number $M_i$ of remaining parameters (equivalently, the dimension of $\mathcal{U}_{\mathrm{loc},i}$) is given by the number of coefficients in the component tensor $u_i$, i.e. $M_i = r_{i-1} \, m \, r_i$. 
If the ranks $r_i,r_{i-1}$ are significantly smaller than $M$, this results in a low-dimensional hence efficiently solvable least-squares problem. Iterating over the component tensors $u_i$ then leads to an efficient scheme for solving high-dimensional least-squares problems with low rank structure.
Basis functions in $\mathcal{U}_{\mathrm{loc},i}$ are obtained from the order 3 tensor $b^{\mathrm{loc}}$ depicted in Figure \ref{fig: graph local basis function}
(note the three open edges). A simple reshape to an order one tensor then yields the desired basis functions, stacked onto each other, i.e. $b^{\mathrm{loc},i}(x) = [b^{\mathrm{loc},i}_1 (x), b^{\mathrm{loc},i}_2(x), \dots, b^{\mathrm{loc},i}_{M_i}(x)]$.

\review{
More precisely, the local basis functions can be identified using the open edges in Figure \ref{fig: graph local basis function} as follows.
Assuming $u_2$ is being optimized, we notice that the tensor $\phi(x_1) \circ u_1$ is a mapping from $\mathbb R \to \mathbb R^{r_1}$, which means that we can identify $r_1$ many one-dimensional functions.
Note that this corresponds to the left part of the tensor picture in Figure \ref{fig: graph local basis function}.
Further, we have that $\phi(x_2)$ is a vector consisting of $m$ one-dimensional functions, which is the middle part of the above tensor picture.
The right part, consisting of the contractions between $\phi(x_2)$, $u_3$, $u_4$, and $\phi(x_4)$, is a set of two-dimensional functions with cardinality $r_2$.
Taking the tensor product of the above functions yields an $r_1  m   r_2$ dimensional function space of four-dimensional functions, which is exactly the span of the local basis functions.}

Further details as well as explicit formulas are given in Appendix \ref{sec:local_basis}.

\begin{figure}[h!]
    \centering
    \begin{tikzpicture}
        \begin{scope}[every node/.style={draw,  fill=white}]
        \node (A1) at (0,0) {$u_1$}; 
        \node (A3) at (2.5,0) {$u_3$}; 
        \node (A4) at (3.75,0) {$u_4$}; 
        
        \node (B1) at (0,-1) {$\phi(x_1)$}; 
        \node (B2) at (1.25,-1) {$\phi(x_2)$}; 
        \node (B3) at (2.5,-1) {$\phi(x_3)$}; 
        \node (B4) at (3.75,-1) {$\phi(x_4)$}; 
        \end{scope}
        \node (A2) at (1.25,0) {}; 
        \node (C0) at (-2,0) {$b^{\mathrm{loc},i}(x)$}; 
        \node (C1) at (-1,0) {$=$}; 
        \begin{scope}[every edge/.style={draw=black,thick}]
        	\path [-] (A1) edge node[midway,left,sloped] [above] {$r_1$} (A2);
        	\path [-] (A2) edge node[midway,left,sloped] [above] {$r_2$} (A3);
        	\path [-] (A3) edge node[midway,left,sloped] [above] {$r_3$} (A4);
        	\path [-] (A1) edge node[midway,left] [right] {$m$} (B1);
        	\path [-] (A2) edge node[midway,left] [right] {$m$} (B2);
        	\path [-] (A3) edge node[midway,left] [right] {$m$} (B3);
        	\path [-] (A4) edge node[midway,left] [right] {$m$} (B4);
        	
        \end{scope}
    \end{tikzpicture}
    \caption{Graphical representation of the local basis functions for $i = 2$.}
    \label{fig: graph local basis function}
\end{figure}
In many situations the terminal condition $g$, defined in \eqref{eq: definition boundary value}, is not part of the ansatz space just defined.
This is always the case if $g$ is not in tensor-product form.
However, as the ambient space $\mathbb{R}^{m^d}$ is linear, $g$ can be straightforwardly added\footnote{We note that the idea of enhancing the ansatz space has been suggested in \cite{zhang2017backward} in the context of linear parametrizations.} to the ansatz space, potentially increasing its dimension to $m^d+1$.
Whenever a component tensor $u_i$ is optimized in the way described above, we simply add $g$ to the set of local basis functions, obtaining as a new basis
\begin{equation}\label{eq:loal_basis}
    b_g^{\mathrm{loc},i} = \{ b_1^{\mathrm{loc},i}, \dots, b_m^{\mathrm{loc},i}, g \},
\end{equation}
only marginally increasing the complexity of the least-squares problem.
In our numerical tests we have noticed substantial improvements using the extension \eqref{eq:loal_basis}.
Incorporating the terminal condition, the representation of the PDE solution takes the form depicted in Figure \ref{fig:valuefun3}, for some $c_g \in \mathbb{R}$.
\begin{figure}[h!]
    \centering
    \begin{tikzpicture}
        \begin{scope}[every node/.style={draw,  fill=white}]
        \node (A1) at (0,0) {$u_1$}; 
        \node (A2) at (1.25,0) {$u_2$}; 
        \node (A3) at (2.5,0) {$u_3$}; 
        \node (A4) at (3.75,0) {$u_4$}; 
        
        \node (B1) at (0,-1) {$\phi(x_1)$}; 
        \node (B2) at (1.25,-1) {$\phi(x_2)$}; 
        \node (B3) at (2.5,-1) {$\phi(x_3)$}; 
        \node (B4) at (3.75,-1) {$\phi(x_4)$}; 
        \end{scope}
        \node (C0) at (-2,0) {$\widehat V(x)$}; 
        \node (C1) at (-1,0) {$=$}; 
        \node (C2) at (5,0) {$+\,\,\, c_g g(x)$}; 
        \begin{scope}[every edge/.style={draw=black,thick}]
        	\path [-] (A1) edge node[midway,left,sloped] [above] {$r_1$} (A2);
        	\path [-] (A2) edge node[midway,left,sloped] [above] {$r_2$} (A3);
        	\path [-] (A3) edge node[midway,left,sloped] [above] {$r_3$} (A4);
        	\path [-] (A1) edge node[midway,left] [right] {$m$} (B1);
        	\path [-] (A2) edge node[midway,left] [right] {$m$} (B2);
        	\path [-] (A3) edge node[midway,left] [right] {$m$} (B3);
        	\path [-] (A4) edge node[midway,left] [right] {$m$} (B4);
        	
        \end{scope}
    \end{tikzpicture}
    \caption{Graphical representation of $\widehat V: \mathbb R^4 \to \mathbb R$.}
    \label{fig:valuefun3}
\end{figure}

Summing up, we briefly state a basic ALS algorithm with our adapted basis $b^{\mathrm{loc},i}$:

\begin{algorithm}[h!]
   \caption{simple ALS algorithm}
   \label{alg:salsa}
\begin{algorithmic}
   \STATE {\bfseries Input:} initial guess $u_1 \circ u_2 \circ \dots \circ u_d$.
   \STATE {\bfseries Output:} result $u_1 \circ u_2 \circ \dots \circ u_d$.
   \REPEAT
   \FOR{$i=1$ {\bfseries to} $d$}
   \STATE identify the local basis functions \eqref{eq:loal_basis}, parametrized by $u_k$, $k \neq j$
   \STATE optimize $u_i$ using the local basis by solving the local least squares problem
   \ENDFOR
   \UNTIL{$noChange$ is $true$}
\end{algorithmic}
\end{algorithm}
The drawback of Algorithm \ref{alg:salsa} is that the ranks of the tensor approximation have to be chosen in advance.
However, there are more involved rank-adaptive versions of the ALS algorithm, providing a convenient way of finding suitable ranks.
In this paper we make use of the rank-adaptive \emph{stable alternating least-squares algorithm (SALSA)} \cite{grasedyck2019stable}.
However, as we will see in Section \ref{sec: numerical examples}, we can in fact oftentimes find good solutions by setting the rank to be $(1, \dots, 1) \in \mathbb N^{d-1}$, enabling highly efficient computations.

By straightforward extensions, adding the terminal condition $g$ to to set of local ansatz functions can similarly be implemented into more advanced, rank adaptive ALS algorithms, which is exactly what we do for our version of SALSA.

\subsection{Handling implicit regression problems}
\label{sec: handling non-linear equations}
The algorithms described in the previous section require the regression problem  to be explicit such as in \eqref{eq: projection_based_optimization}.
In contrast, the optimization in \eqref{eq: implicit scheme} is of implicit type, as $\widehat{h}_n$ contains the unknown $\widehat{V}_n$.
In order to solve \eqref{eq: implicit scheme}, we therefore choose an initial guess $\widehat V_{n}^0$ and iterate the optimization of
\begin{multline}\label{eq:fixed-point}
    \E[ (\widehat{V}_n^{k+1}(\widehat{X}_n) - h(\widehat X_n, t_n, \widehat Y_n^{k}, \widehat Z_n^{k}) \Delta t +  \\
    \widehat{Z}_n^{k} \cdot \xi_{n+1}\sqrt{\Delta t}  - \widehat{V}_{n+1}(\widehat{X}_{n+1}))^2]
\end{multline}
with respect to  $\widehat V_n^{k+1}$ until convergence (see Appendix \ref{sec:implementation} for a discussion of appropriate stopping criteria). In the above display, $\widehat{Y}_n^k = \widehat{V}^k_n(\widehat{X}_n)$ and $\widehat{Z}_n^k = (\sigma^\top \nabla \widehat{V}_n^k)(\widehat{X}_n)$ are computed according to \eqref{eq: def Y Z}.
For theoretical foundation, we guarantee convergence of the proposed scheme when the step size $\Delta t$ is  small enough.
\begin{theorem}\label{thm:fixed_point}
Assume that $\mathcal{U} \subset L^2(\Omega) \cap C_b^\infty(\Omega)$ is a finite dimensional linear subspace, that $\sigma(x,t)$ is
nondegenerate for all $(x,t) \in [0,T] \times \mathbb{R}^d$, and that $h$ is globally Lipschitz continuous in the last two arguments. Then there exists $\delta > 0$ such that the iteration \eqref{eq:fixed-point} converges for all $\Delta t \in (0,\delta)$.
\end{theorem}
\begin{proof}
See Appendix \ref{app:proof}.
\end{proof}

\begin{remark}\label{rem:regularization}
In order to ensure the boundedness assumption in Theorem \ref{thm:fixed_point} and to stabilize the computation we add a regularization term involving the Frobenius norm of the coefficient tensor to the objective in  \eqref{eq:fixed-point}.
Choosing an orthonormal basis we can then relate the Frobenius norm to the associated norm in the function space by Parseval's identity.
In our numerical tests we set our one-dimensional ansatz functions to be $H^2(a, b)$-orthonormal
\footnote{Here, $H^2(a,b)$ refers to the second-order Sobolev space, see \cite{sickel2009tensor}.},
\review{
where $a$ and $b$ are set to be approximately equal to the minimum and maximum of the samples $\widehat X_n$, respectively.
In Appendix \ref{app: HJB} we state the exact choices of $a$ and $b$ for the individual numerical tests.}
The corresponding tensor space $(H^2(a,b))^{\otimes d} = H^2_{\text{mix}}([a, b])^d$ can be shown to be continuously embedded in $W^{1, \infty}(\Omega)$, guaranteeing boundedness of the approximations and their derivatives \cite{sickel2009tensor}. 
\end{remark}

\begin{remark}[Parameter initializations] 
\label{rem: parameter initializations}
Since we expect $V(\cdot, t_n)$ to be close to $V(\cdot, t_{n+1})$ for any $n \in \{0, \dots, N-1\}$, we initialize the parameters of $\widehat{V}^0_n$ as those obtained for $\widehat{V}_{n+1}$ identified in the  preceding time step.
\end{remark}

\review{Clearly, the iterative optimization of \eqref{eq:fixed-point} is computationally more costly than the explicit scheme described in Section \ref{sec: optimization on TT manifold} that relies on a single optimization of the type \eqref{eq:lin_least_squares} per time step. However, implicit schemes typically ensure improved convergence orders as well as robustness \cite{kloeden1992stochastic} and therefore hold the promise of more accurate approximations (see Section \ref{sec: numerical examples} for experimental confirmation).
We note that the NN based approaches considered as baselines in Section \ref{sec: numerical examples} perform gradient descent for both the explicit and implicit schemes and therefore no significant differences in the corresponding runtimes are expected.} For convenience, we summarize the developed methods in Algorithm \ref{alg2}.

\begin{algorithm}[h!]
   \caption{PDE approximation}
\label{alg2}
\begin{algorithmic}
   \STATE {\bfseries Input:} initial parametric choice for the functions $\widehat{V}_n$ for $n \in \{0, \dots, N-1 \}$
   \STATE {\bfseries Output:} approximation of $V(\cdot, t_n) \approx \widehat{V}_n$ along the trajectories for $n \in \{0, \dots, N-1 \}$
   \STATE Simulate $K$ samples of the discretized SDE \eqref{eq: discrete SDE}.   
   \STATE Choose $\widehat{V}_N = g$.
   \FOR{$n = N - 1$ {\bfseries to} $0$}
   \STATE approximate either \eqref{eq: projection_based_optimization} or \eqref{eq: implicit scheme} (both depending on $\widehat{V}_{n+1}$) using Monte Carlo
   \STATE minimize this quantity (explicitly or by iterative schemes)
   \STATE set $\widehat{V}_n$ to be the minimizer
   \ENDFOR
\end{algorithmic}
\end{algorithm}

\section{Numerical examples}
\label{sec: numerical examples}

In this section we consider some examples of high-dimensional PDEs that have been addressed in recent articles and treat them as benchmark problems in order to compare against our algorithms with respect to approximation accuracy and computation time. We refer to Appendix \ref{app: implementation details} for implementation details and to Appendix \ref{app: further numerical examples and details} for additional experiments.

\subsection{Hamilton-Jacobi-Bellman equation}
\label{sec: HJB equation}

The Hamilton-Jacobi-Bellman equation (HJB) is a PDE for the so-called value function that represents the minimal cost-to-go in stochastic optimal control problems from which the optimal control policy can be deduced. As suggested in \cite{weinan2017deep}, we consider the HJB equation
\begin{subequations}
\begin{align}
    \left(\partial_t + \Delta \right)V(x, t) - |\nabla V(x, t)|^2 &= 0, \\
    V(x, T) &= g(x),
\end{align}
\end{subequations}
with $g(x) = \log\left(\frac{1}{2} + \frac{1}{2}|x|^2 \right)$, leading to
\begin{equation}
    b = \mathbf{0},\quad \sigma = \sqrt{2} \, \mathrm{Id}_{d \times d},\quad h(x, s, y, z) = -\frac{1}{2}|z|^2
\end{equation}
in terms of the notation established in Section \ref{sec: solving PDEs via BSDEs}.
One appealing property of this equation is that (up to Monte Carlo approximation)  a reference solution is available: 
\begin{equation}
\label{eq: HJB reference solution}
    V(x, t) = -\log \E \left[e^{-g(x + \sqrt{T- t}\sigma \xi)} \right],
\end{equation}
where $\xi \sim \mathcal{N}(\mathbf{0}, \mathrm{Id}_{d\times d})$ is a normally distributed random variable (see Appendix \ref{app: HJB} for further details).

In our experiments we consider $d = 100, T = 1, \Delta t = 0.01, x_0 = (0, \dots, 0)^\top$ and $K = 2000$ samples. In Table \ref{tbl: HJB d = 100} we compare the explicit scheme stated in \eqref{eq: projection_based_optimization} with the implicit scheme from \eqref{eq: implicit scheme}, once with TTs and once with NNs.
For the tensor trains we try different polynomial degrees, and it turns out that choosing constant ansatz functions is the best choice, while fixing the rank to be $1$.
For the NNs we use a DenseNet like architecture with $4$ hidden layers (all the details can be found in Appendices \ref{app: implementation details} and \ref{app: further numerical examples and details}). 

We display the approximated solutions at $(x_0, 0)$, the corresponding relative errors $\left|\frac{\widehat{V}_n(x_0) - V_\mathrm{ref}(x_0, 0)}{V_\mathrm{ref}(x_0, 0)}\right|$ with $V_\mathrm{ref}(x_0, 0) = 4.589992$ being provided in \cite{weinan2017deep}, their computation times, as well as PDE and reference losses, which are specified in Appendix \ref{app: implementation details}. We can see that the TT approximation is both more accurate and much faster than the NN-based approaches, improving also on the results in \cite{weinan2017deep, beck2019deep}. As it turns out that the explicit scheme for NNs is worse \review{in terms of accuracy} than its implicit counterpart \review{in all our experiments}, but takes a very similar amount of computation time we will omit reporting it for the remaining experiments. In Figures \ref{fig: HJB trajectories, d = 10} and \ref{fig: HJB trajectories} we plot the reference solutions computed by \eqref{eq: HJB reference solution} along two trajectories of the discrete forward process \eqref{eq: discrete SDE} in dimensions $d=10$ and $d=100$ and compare to the implicit TT and NN-based approximations. We can see that the TT approximations perform particularly well in the higher dimensional case $d = 100$.

\begin{table}[h!]
\centering
\begin{tabular}{c | c c c c }
& $\text{TT}_\text{impl}$ & $\text{TT}_\text{expl}$  & $\text{NN}_\text{impl}$& $\text{NN}_\text{expl}$ \\
 \hline
$\widehat{V}_0(x_0)$ & $4.5903$ & $4.5909$  & $4.5822$& $4.4961$ \\
relative error & $5.90 \text{e}^{-5}$ & $3.17 \text{e}^{-4}$& $1.71 \text{e}^{-3}$& $2.05 \text{e}^{-2}$\\

reference loss & $3.55\text{e}^{-4}$ & $5.74\text{e}^{-4}$ & $4.23\text{e}^{-3}$ & $1.91\text{e}^{-2}$ \\
PDE loss & $1.99 \text{e}^{-3}$ & $3.61\text{e}^{-3}$ & $90.89$ & $91.12$ \\
comp. time & $41$ & $25$ & $44712$ & $25178$ \\
\end{tabular}
\caption{Comparison of approximation results for the HJB equation in $d=100$.}
\label{tbl: HJB d = 100}
\end{table}

\begin{figure}[h!]
\vskip 0.2in
\begin{center}
\centerline{\includegraphics[width=1.0\columnwidth]{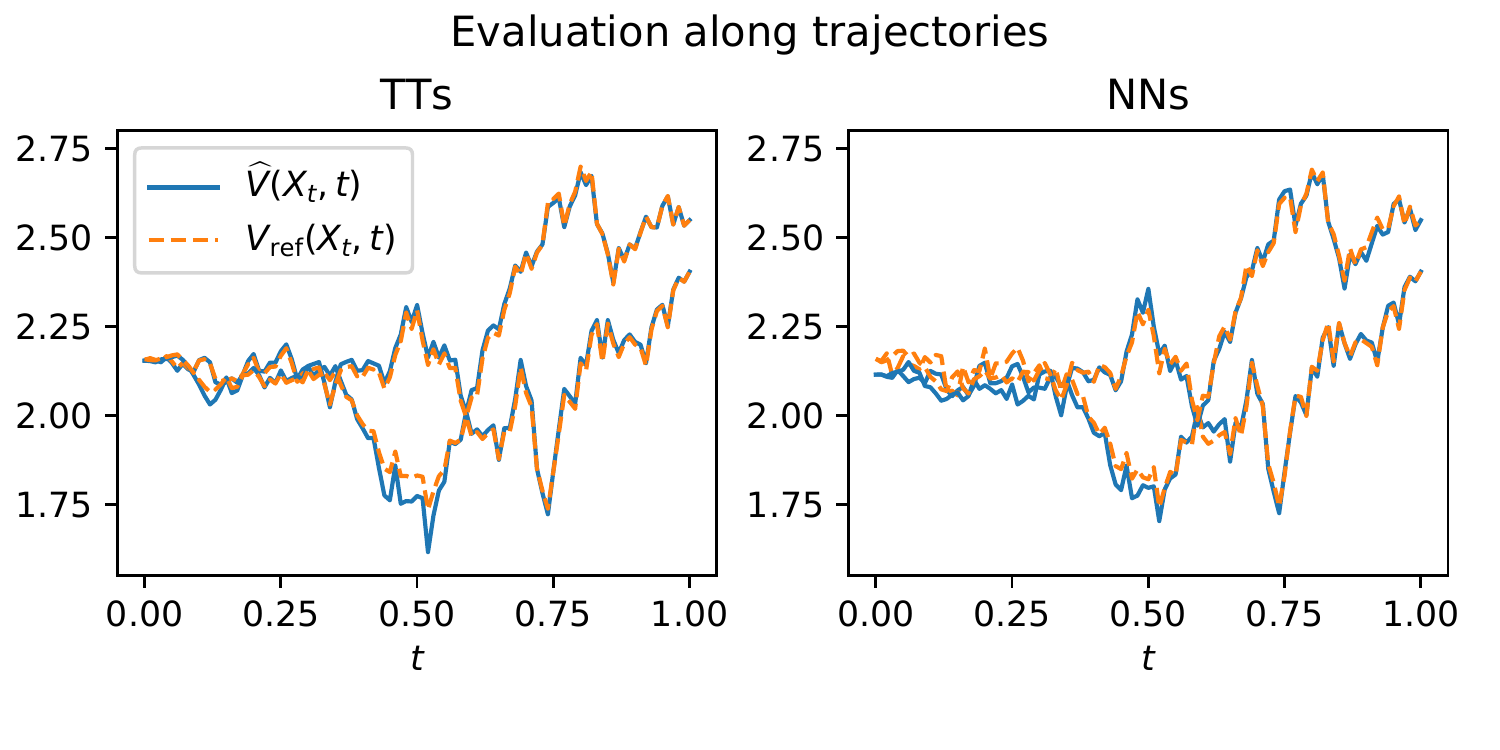}}
\caption{Reference solutions compared with implicit TT and NN approximations along two trajectories in $d=10$.}
\label{fig: HJB trajectories, d = 10}
\end{center}
\vskip -0.2in
\end{figure}

\begin{figure}[h!]
\vskip 0.2in
\begin{center}
\centerline{\includegraphics[width=1.0\columnwidth]{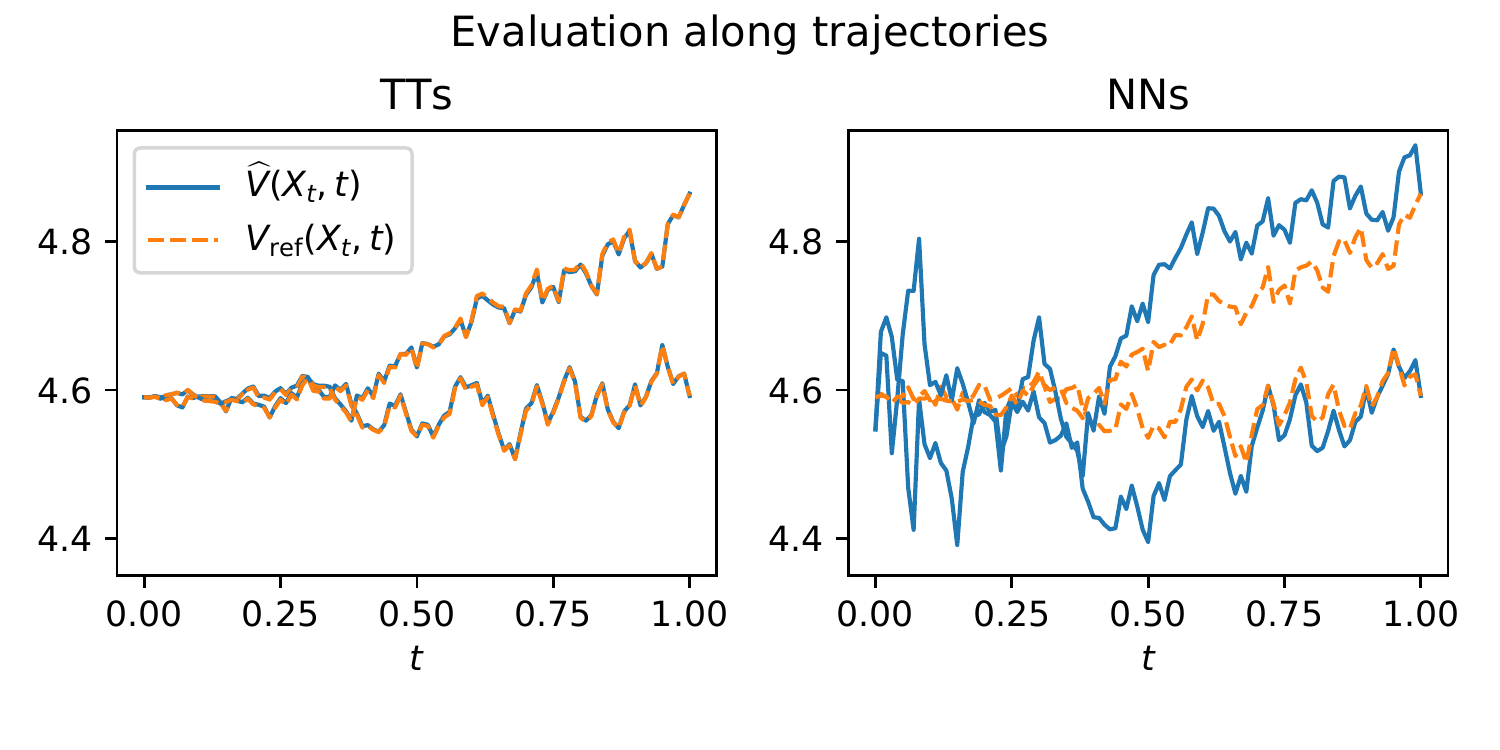}}
\caption{Reference solutions compared with implicit TT and NN approximations along two trajectories in $d=100$.}
\label{fig: HJB trajectories}
\end{center}
\vskip -0.2in
\end{figure}

In Figure \ref{fig: HJB mean relative error} we plot the mean relative error over time, as defined in Appendix \ref{app: implementation details}, indicating that both schemes are stable and where again the implicit TT scheme yields better results than the NN scheme.

\begin{figure}[h!]
\vskip 0.2in
\begin{center}
\centerline{\includegraphics[width=.7\columnwidth]{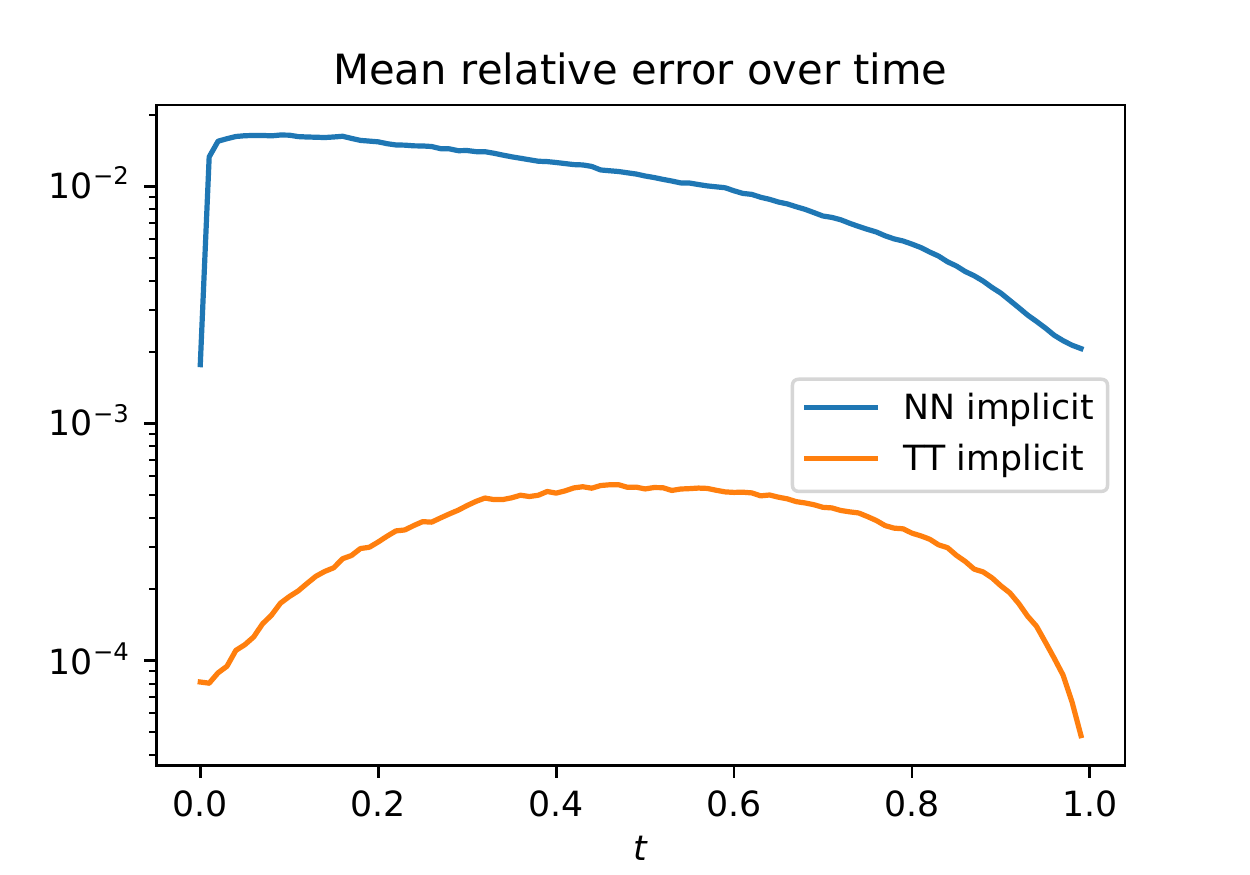}}
\caption{Mean relative error for TT and NN attempts.}
\label{fig: HJB mean relative error}
\end{center}
\vskip -0.2in
\end{figure}

\review{
The accuracy of the TT approximations is surprising given that the ansatz functions are constant in space.  We further investigate this behavior in Table \ref{tab: HJB TTs different poly degress in differen dimensions} and observe that the required polynomial degree decreases with increasing dimension. While similar ``blessings of dimensionality'' have been reported and discussed (see, for instance, Figure 3 in  \cite{bayer2021pricing} and Section 1.3 in \cite{khoromskij2012tensors}), a thorough theoretical understanding is still lacking. To guide intuition, we would like to point out that the phenomenon that high-dimensional systems become in some sense simpler is well known from the theory of interacting particle systems (``propagation of chaos'', see \cite{sznitman1991topics}): In various scenarios, the joint distribution of a large number of particles tends to approximately factorize as the number of particles increases (that is, as the dimensionality of the joint state space grows large). It is plausible that similar approximate factorizations are relevant for high-dimensional PDEs and that tensor methods are useful (i) to detect this effect and (ii) to exploit it. In this experiment, the black-box nature of neural networks does not appear to reveal such properties.} 

\begin{table}[h]
    \centering
    \begin{tabular}{c|c c c c c}
   $d$ & \multicolumn{5}{c}{\text{Polynomial degree}} \\
    & $0$ & $1$ & $2$ & $3$ & $4$  \\
 \hline
  $1$  & $3.62\text{e}^{-1}$ & $3.60\text{e}^{-1}$ & $2.47\text{e}^{-3}$ & $3.86\text{e}^{-4}$ & $4.27\text{e}^{-2}$ \\
  $2$  & $1.03\text{e}^{-1}$ & $1.02\text{e}^{-1}$ & $1.87\text{e}^{-2}$ & $1.79\text{e}^{-2}$ & $1.79\text{e}^{-2}$ \\
  $5$  & $1.55\text{e}^{-2}$ & $1.54\text{e}^{-2}$ & $1.03\text{e}^{-3}$ & $9.52\text{e}^{-4}$ & $1.96\text{e}^{-2}$ \\
  $10$ & $2.84\text{e}^{-3}$ & $2.86\text{e}^{-3}$ & $1.37\text{e}^{-3}$ & $1.34\text{e}^{-3}$ & $1.10\text{e}^{-1}$ \\
  $50$ & $1.17\text{e}^{-4}$ & $1.29\text{e}^{-4}$ & $2.79\text{e}^{-4}$ & $3.35\text{e}^{-4}$ & $6.96\text{e}^{-5}$ \\
  $100$ & $5.90\text{e}^{-5}$ & $4.99\text{e}^{-5}$ & $8.65\text{e}^{-5}$ & $1.23\text{e}^{-4}$ & $3.62\text{e}^{-5}$ \\
         
    \end{tabular}
    \caption{Relative errors of the TT approximations $\widehat{V}_n(x_0)$ for different dimensions and polynomial degrees.}
    \label{tab: HJB TTs different poly degress in differen dimensions}
\end{table}

\subsection{HJB with double-well dynamics}
\label{sec: HJB double well}

In another example we consider again an HJB equation, however this time making the drift in the dynamics nonlinear, as suggested in \cite{nusken2020solving}. The PDE becomes
\begin{subequations}
\begin{align}
    \left(\partial_t + L \right)V(x, t) - \frac{1}{2}|(\sigma^\top \nabla V)(x, t)|^2 &= 0, \\
    V(x, T) &= g(x),
\end{align}
\end{subequations}
with $L$ as in \eqref{eq: infinitesimal generator}, where now the drift is given as the gradient of the double-well potential
\begin{equation}
    b = -\nabla \Psi, \qquad \Psi(x) = \sum_{i,j=1}^d C_{ij}(x_i^2 - 1)(x_j^2 - 1)
\end{equation}
and the terminal condition is $g(x) = \sum_{i=1}^d \nu_i(x_i - 1)^2$ for $\nu_i > 0$. Similarly as before a reference solution is available,
\begin{equation}
\label{eq: HJB double well reference solution}
    V(x, t) = -\log\E\left[e^{-g(X_T)} \Big| X_t = x \right],
\end{equation}
where $X_t$ is the forward diffusion as specified in \eqref{eq: fordward SDE} (see again Appendix \ref{app: HJB} for details).

First, we consider diagonal matrices $C = 0.1 \, \mathrm{Id}_{d \times d}, \sigma = \sqrt{2} \,\mathrm{Id}_{d \times d}$, implying that the dimensions do not interact, and take $T = 0.5, d = 50, \Delta t = 0.01, K = 2000, \nu_i = 0.05$. 
\review{We set the TT-rank to $2$, use polynomial degree $3$ and refer to Appendix \ref{app: further numerical examples and details} for further details on the TT and NN configurations}. Since in the solution of the PDE the dimensions do not interact either, we can compute a reference solution with finite differences. In Table \ref{tbl: HJB double well diagonal} we see that the TT and NN approximations are compatible with TTs having an advantage in computational time.

\begin{table}[h!]
\centering
\begin{tabular}{c | c c c c }
& $\text{TT}_\text{impl}$  & $\text{TT}_{\text{expl}}$ &  $\text{NN}_\text{impl}$ \\
 \hline
$\widehat{V}_0(x_0)$ & $\review{9.6876}$ &  $\review{9.6865}$ &  $9.6942$  \\
relative error & \review{$1.41\text{e}^{-3}$} & \review{$1.53\text{e}^{-3}$} &  $7.27\text{e}^{-4}$ \\
reference loss & \review{$1.36\text{e}^{-3}$} & \review{$3.25\text{e}^{-3}$} &  $4.25\text{e}^{-3}$  \\
PDE loss & $\review{3.62\text{e}^{-2}}$ &  $\review{11.48}$& $2.66\text{e}^{-1}$ \\
computation time & $\review{95}$ & $\review{16}$ &  $1987$ \\
\end{tabular}
\caption{Approximation results for the HJB equation with non-interacting double well potential in $d = 50$.}
\label{tbl: HJB double well diagonal}
\end{table}

Let us now consider a non-diagonal matrix $C = \mathrm{Id}_{d \times d} + (\xi_{ij})$, where $\xi_{ij} \sim \mathcal{N}(0, \review{0.1})$ are sampled once at the beginning of the experiment and further choose $\sigma = \sqrt{2} \,\mathrm{Id}_{d \times d}, \nu_i = 0.5, T = 0.3$.
We aim at the solution at $x_0 = (-1, \dots, -1)^\top$ and  compute a reference solution with \eqref{eq: HJB double well reference solution} using $10^7$ samples. We see in Table \ref{tbl: HJB double well nondiagonal} that TTs are much faster than NNs, while yielding a similar performance.
\review{Note that due to the non-diagonality of $C$ it is expected that the TTs are of rank larger than $2$.
For the explicit case we do not cap the ranks of the TT and the rank-adaptive solver finds ranks of mostly $4$ and never larger than $6$.
Motivated by these results we cap the ranks at $r_i \leq 6$ in the implicit case and indeed they are obtained for nearly every dimension, as seen from the ranks below,
\[[5, 6, 6, 6, 6, 6, 6, 6, 6, 6, 6, 6, 6, 6, 6, 6, 6, 6, 5].  \]
The results were obtained with polynomial degree $7$.
}

\begin{table}[h!]
\centering
\begin{tabular}{c | c c c c }
& $\text{TT}_\text{impl}$  & $\text{TT}_\text{expl}$  & $\text{NN}_\text{impl}$ \\
 \hline
$\widehat{V}_0(x_0)$ & $35.015$ & $34.756$ &  $34.917$ \\
relative error & $1.52\text{e}^{-3}$&  $2.82\text{e}^{-3}$ &    $4.24\text{e}^{-3}$ \\
reference loss & $1.30\text{e}^{-2}$ &  $1.59\text{e}^{-2}$ &  $6.38\text{e}^{-2}$  \\
PDE loss & $79.9$ &  $341$  &   $170.64$ \\
computation time & $460$ & $15$ &  $16991$ \\
\end{tabular}
\caption{Approximation results for the HJB equation with interacting double well potential in $d = 20$.}
\label{tbl: HJB double well nondiagonal}
\end{table}

\subsection{Cox–Ingersoll–Ross model}\label{sec:cir}
Our last example is taken from financial mathematics. As suggested in \cite{jiang2021convergence} we consider a bond price in a multidimensional Cox–Ingersoll–Ross (CIR) model, see also \cite{hyndman2007forward, alfonsi2015affine}. The underlying PDE is specified as
\begin{multline}
    \partial_t V(x, t) + \frac 1 2 \sum_{i,j = 1}^d \sqrt{ x_i x_j} \gamma_i \gamma_j \partial_{x_i} \partial_{x_j} V(x, t) \\ 
    + \sum_{i = 1}^d a_i (b_i - x_i) \partial_{x_i} V(x, t)  - \left(\max_{1 \leq i \leq d} x_i \right) V(x, t) = 0.
\end{multline}
Here, $a_i, b_i, \gamma_i \in [0, 1]$ are uniformly sampled at the beginning of the experiment and $V(T, x) = 1$. We set $d = 100$.

We aim to estimate the bond price at the initial condition $x_0 = (1, \dots, 1)^\top$. As there is no reference solution known, we rely on the PDE loss to compare our results. Table \ref{tbl: cir results} shows that all three approaches yield similar results, while having a rather small PDE loss.
\review{For this test it is again sufficient to set the TT-rank to $1$ and the polynomial degree to $3$.}
The TT approximations seem to be slightly better and we note that the explicit TT scheme is again much faster.

\begin{table}[h!]
\centering

\begin{tabular}{c | c c c }
& $\text{TT}_\text{impl}$ &  $\text{TT}_\text{expl}$ &  $\text{NN}_\text{impl}$ \\
 \hline
$\widehat{V}_0(x_0)$ & $0.312$ & $0.306$ &$0.31087$ \\
PDE loss & $5.06\text{e}^{-4}$ & $5.04\text{e}^{-4}$ & $7.57\text{e}^{-3}$ \\
computation time & $5281$& $197$ & $9573$ \\
\end{tabular}
\caption{$K = 1000$, $d = 100$, $x_0 = [1,1,\dots, 1]$}
\label{tbl: cir results}
\end{table}

In Table  \ref{tab: CIR TTs different poly degress in differen dimensions} we compare the PDE loss using different polynomial degrees for the TT ansatz function and see that we do not get any improvements with polynomials of degree larger than $1$.

\begin{figure}[h!]
\vskip 0.2in
\begin{center}
\centerline{\includegraphics[width=1.0\columnwidth]{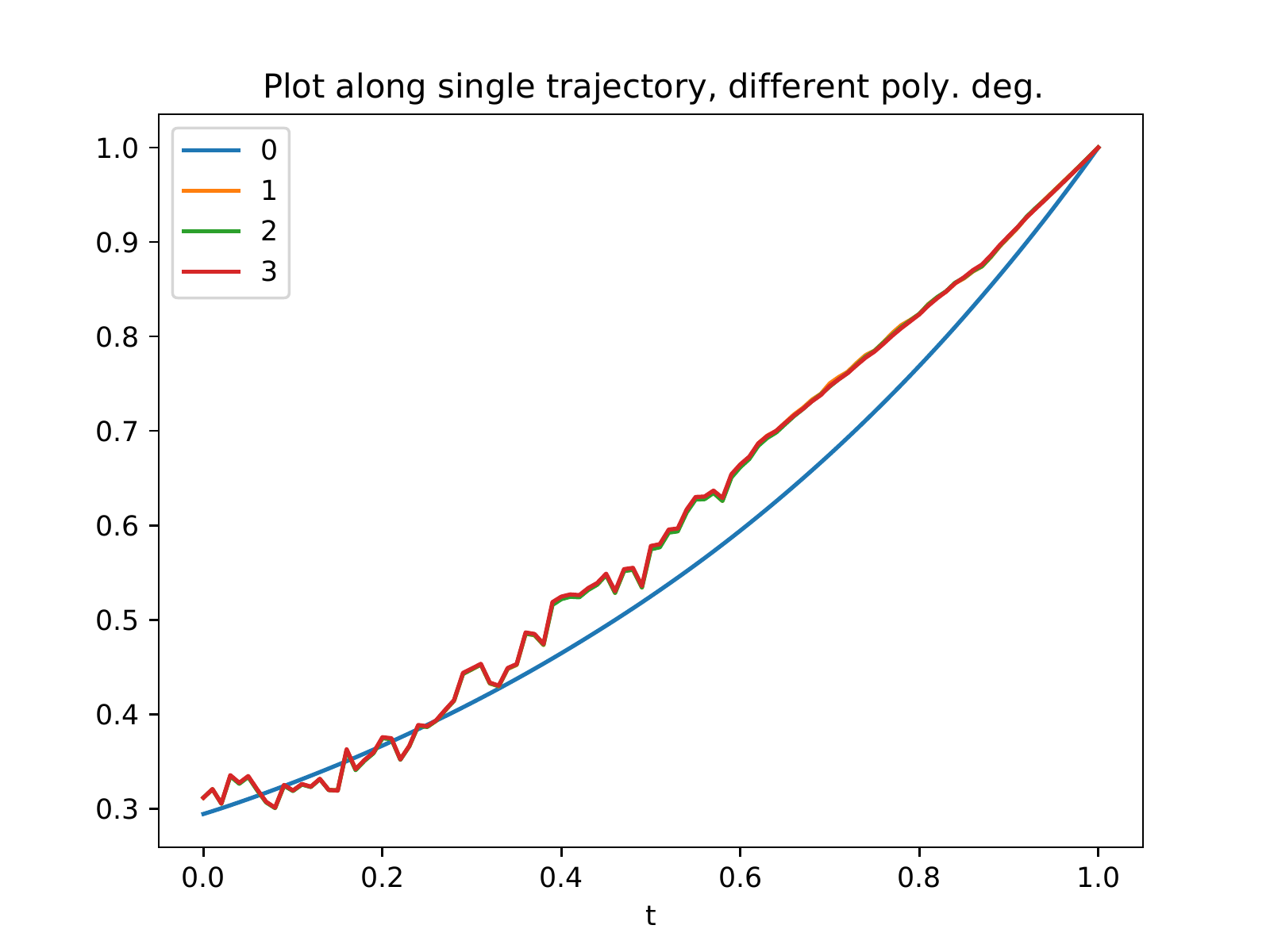}}
\caption{Reference trajectory for different polynomial degrees.}
\label{fig: cir trajectories}
\end{center}
\vskip -0.2in
\end{figure}

\begin{table}[h]
    \centering
    \begin{tabular}{c|c c c c}
  & \multicolumn{4}{c}{\text{Polynom. degree}} \\
    & $0$ & $1$ & $2$ & $3$ \\
 \hline
$\widehat{V}_0(x_0)$ & $0.294$ & $0.312$ & $0.312$ & $0.312$ \\
PDE loss & $9.04\text{e}^{-2}$ & $7.80\text{e}^{-4}$ & $1.05\text{e}^{-3}$ & $5.06\text{e}^{-4}$  \\
comp. time & $110$ & $3609$ & $4219$ & $5281$ \\
    \end{tabular}
    \caption{PDE loss and computation time for TTs with different polynomial degrees}
    \label{tab: CIR TTs different poly degress in differen dimensions}
\end{table}
Noticing the similarity between the results for polynomial degrees $1$, $2$, and $3$, we further investigate by computing the value function along a sample trajectory in Figure \ref{fig: cir trajectories}, where we see that indeed the approximations with those polynomial degrees are indistinguishable.

\review{
\section{Conclusions and outlook}
In this paper, we have developed tensor train based approaches towards solving high-dimensional parabolic PDEs, relying on reformulations in terms of BSDEs. For the discretization of the latter, we have considered both explicit and implicit schemes, allowing for a trade-off between approximation accuracy and computational cost. Notably, the tensor train format specifically allows us to take advantage of the additional structure inherent in least-squares based formulations, particularly in the  explicit case. 
} 

\review{
More elaborate numerical treatments for BSDEs (involving, for instance, multi-step and/or higher-order discretizations) have been put forward in the literature \cite{chassagneux2014linear,crisan2014second,macris2020solving}. Combining these with tensor based methods remains a challenging and interesting topic for future research. 
Finally, we believe that the ``blessing of dimensionality'' observed in Section \ref{sec: HJB equation} deserves a mathematically rigorous explanation; progress in this direction may further inform the design of scalable schemes for high-dimensional PDEs. 
}

 \textbf{Acknowledgements}
 This research has been partially funded by 
 Deutsche Forschungsgemeinschaft (DFG) through the grant 
 CRC 1114 \lq Scaling Cascades in Complex Systems\rq \,(projects A02 and A05, project number 235221301).
 L. S. acknowledges support from the Research Training Group \lq Differential Equation- and Data-driven Models in Life Sciences and Fluid Dynamics: An Interdisciplinary Research Training Group (DAEDALUS)\rq  (GRK 2433) funded by Deutsche Forschungsgemeinschaft (DFG).
 We would like to thank Reinhold Schneider for giving valuable input and for sharing his broad insight in tensor methods and optimization.

\bibliography{references}
\bibliographystyle{icml2020}

\newpage

\appendix

\section{Graphical notation for tensor trains}\label{sect:graphical_notation}
In this section we provide some further material on tensor networks and their graphic notation.
Let us start by noting that a vector $x \in \mathbb R^n$ can be interpreted as a tensor.
\begin{figure}[h]
    \centering
        \begin{tikzpicture}
        \begin{scope}[every node/.style={draw,  fill=white}]
                
                \node (A1) at (0,0) {$x$}; 
            \end{scope}
            \begin{scope}[every edge/.style={draw=black,thick}]
            	\path [-] (A1) edge node[midway,left] [above] {$n$} (0:1);
            \end{scope}
        \end{tikzpicture}
        \end{figure}

\noindent In the graphic representation contractions between indices are denoted by a line between the tensors.
Below we contract a tensor $A \in \mathbb R^{n \times m}$ and $x \in \mathbb R^n$, which results in an element of $\mathbb R^m$, representing the usual matrix-vector product.
        \begin{figure}[h]
        \centering
        \begin{tikzpicture}
        \begin{scope}[every node/.style={draw,  fill=white}]
                
                \node (A1) at (0,0) {$x$}; 
               	\node (A2) at (1, 0) {$A$};  	
            \end{scope}
            \begin{scope}[every edge/.style={draw=black,thick}]
            	\path [-] (A1) edge node[midway,left] [above] {$n$} (A2);
            	\path [-] (A2) edge node[midway,left] [above] {$m$} (0:2.5);
            \end{scope}
        \end{tikzpicture}
        \end{figure}
        
\noindent
In Figure \ref{fig: order 3 tensor} an order $3$ tensor $B \in \mathbb R^{n_1 \times n_2 \times n_3}$ is represented with three lines, not connected to any other tensor.
        \begin{figure}[h]
            \centering
        \begin{tikzpicture}
        \begin{scope}[every node/.style={draw,  fill=white}]
                
                \node (A1) at (0,0) {$\mathbf B$}; 
            \end{scope}
            \begin{scope}[every edge/.style={draw=black,thick}]
            	\path [-] (A1) edge node[midway,left,sloped] [above] {$n_1$} (0:1);
            	\path [-] (A1) edge node[midway,left,sloped] [above] {$n_2$} (-60:-1);
				\path [-] (A1) edge node[midway,left,sloped] [above] {$n_3$} (60:-1);
            \end{scope}
        \end{tikzpicture}
    \caption{Graphical notation of simple tensors and tensor networks}
    \label{fig: order 3 tensor}
\end{figure}
\noindent %
As another example, we can write the compact singular value decomposition in matrix form as $A = U \Sigma V$, with $U \in \R^{n, r}, \Sigma \in \R^{r, r}, V \in \R^{r, m}$, which we represent as a tensor network in Figure \ref{fig: SVD}.
        \begin{figure}[h]
            \centering
    \begin{tikzpicture}
        \begin{scope}[every node/.style={draw,  fill=white}]
        \node (a1) at (0,0) {$U$}; 
        \node (a2) at (1.25,0) {$\Sigma$}; 
        \node (a3) at (2.5,0) {$V$}; 
        \node (A0) at (-3,0) {$A$}; 
        \end{scope}
        \node (B2) at (-1.5,0) {$=$}; 
        \begin{scope}[every edge/.style={draw=black,thick}]
        	\path [-] (a1) edge node[midway,left,sloped] [above] {$r$} (a2);
        	\path [-] (a2) edge node[midway,left,sloped] [above] {$r$} (a3);

        	\path [-] (a1) edge node[midway,left] [above] {$n$} +(180:1);
        	\path [-] (a3) edge node[midway,left] [above] {$m$} +(0:1);
        	
        	\path [-] (A0) edge node[midway,left] [above] {$n$} +(180:1);
        	\path [-] (A0) edge node[midway,left] [above] {$m$} +(0:1);
        \end{scope}
    \end{tikzpicture}
    \caption{Graphical notation of simple tensors and tensor networks.}
    \label{fig: SVD}
\end{figure}

\subsection{The local basis functions}\label{sec:local_basis}
\review{Following the inexact description of the local basis functions we now give a precise formula.}
When optimizing the $k$-th component tensor, the local basis functions are given by setting $ 1 \leq j_{k-1} \leq r_{k-1}$, $1 \leq i_k \leq m$, and $1 \leq j_k \leq r_k$ within the following formula:

\begin{align}
\begin{split}
    &b_{j_{k-1}, i_k, j_k}(x) = \\ 
    &\quad\Bigg( \sum_{i_1, \dots, i_{k-1}}^{m, \dots, m} \sum_{j_1, \dots, j_{k-2}}^{r_1, \dots, r_{k-2}} u_1[i_1, j_1] \dots u_{k-1}[j_{k-2}, i_{k-1}, j_{k-1}] \\
    &\quad\qquad\phi(x_1)[i_1] \dots \phi(x_{k-1})[i_{k-1}] \Bigg) \phi(x_k)[i_k]\\
    &\quad \Bigg( \sum_{i_{k+1}, \dots, i_{d}}^{m, \dots, m} \sum_{j_{k}, \dots, j_{d-1}}^{r_{k}, \dots, r_{d-1}} u_{k+1}[j_k, i_{k+1}, j_{k+1}] \dots u_{d}[j_{d-1}, i_{d}] \\
    &\quad\qquad \phi(x_{k+1})[i_{k+1}] \dots \phi(x_{d})[i_{d}] \Bigg).
\end{split}
\end{align}
Note that in the above formula, every index except $j_{k-1}$, $i_k$ and $j_{k}$ is contracted, leaving an order three tensor.
A simple reshape into one index then yields the local basis functions as used in this paper.
\section{Proof of Theorem \ref{thm:fixed_point}}
\label{app:proof}

\begin{proof}[Proof of Theorem \ref{thm:fixed_point}]
In this proof, we denote the underlying probability measure by $\mathbb{P}$, and the corresponding Hilbert space of random variables with finite second moments by $L^2(\mathbb{P})$. We define the linear subspace $\widetilde{\mathcal{U}} \subset L^2(\mathbb{P})$ by 
\begin{equation}
\widetilde{\mathcal{U}} = \left\{ f(\widehat{X}_n) : f \in \mathcal{U}  \right\},    
\end{equation}
noting that $\widetilde{\mathcal{U}}$ is finite-dimensional by the assumption on $\mathcal{U}$, hence closed. The corresponding $L^2(\mathbb{P})$-orthogonal projection onto  $\widetilde{\mathcal{U}}$ will be denoted by $\Pi_{\widetilde{\mathcal{U}}}$. By the nondegeneracy of $\sigma$, the law of $\widehat{X}_n$ has full support on $\Omega$, and so $\Vert \cdot \Vert_{L^2(\mathbb{P})}$ is indeed a norm on $\widetilde{\mathcal{U}}$.  Since $\widetilde{\mathcal{U}}$ is finite-dimensional, the linear operators
\begin{equation}
\widetilde{\mathcal{U}} \ni f(\widehat{X}_n) \mapsto \frac{\partial f}{\partial x_i}(\widehat{X}_n) \in L^2(\mathbb{P})     
\end{equation}
are bounded, and consequently there exists a constant $C_1>0$ such that \begin{equation}
\label{eq:der estimate}
\left \Vert \frac{\partial f}{\partial x_i}(\widehat{X}_n) \right \Vert_{L^2(\mathbb{P})} \le C_1 \left\Vert f(\widehat{X}_n) \right\Vert_{L^2(\mathbb{P})},      
\end{equation}
for all $i=1,\ldots,d$ and $f \in \mathcal{U}$. 
Furthermore, there exists a constant $C_2 > 0$ such that
\begin{equation}
\label{eq:moment estimate}
    \nonumber 
    \mathbb{E} \left[ f^4(\widehat{X}_n)\right]^{1/4} :=
    \left \Vert f(\widehat{X}_N) \right\Vert_{L^4(\mathbb{P})} \le C_2 \left \Vert f (\widehat{X}_n)\right\Vert_{L^2(\mathbb{P})},  
\end{equation}
for all $f \in \mathcal{U}$,
again by the finite-dimensionality of $\widetilde{\mathcal{U}}$ and the fact that on finite dimensional vector spaces, all norms are equivalent.
By standard results on orthogonal projections, the solution to the iteration \eqref{eq:fixed-point} is given by
\begin{subequations}
\begin{align}
V_n^{k+1}(\widehat{X}_n) =  \Pi_{\widetilde{\mathcal{U}}} \big[ 
- h(\widehat X_n, t_n, \widehat Y_n^{k}, \widehat Z_n^{k}) \Delta t + 
\nonumber
\\
    \widehat{Z}_n^{k} \cdot \xi_{n+1}\sqrt{\Delta t}  - \widehat{V}_{n+1}(\widehat{X}_{n+1})  \big].
    \nonumber
\end{align}
\end{subequations}
We now consider the map $\Psi: \widetilde{\mathcal{U}} \rightarrow \widetilde{\mathcal{U}}$ defined by
\begin{subequations}
\nonumber
\begin{align}
f(\widehat{X}_n) \mapsto 
\Pi_{\widetilde{\mathcal{U}}} \big[ 
- h(\widehat X_n, t_n, f(\widehat{X}_n), \sigma^\top \nabla f(\widehat{X}_n)) \Delta t +  \\
    \sigma^\top \nabla f(\widehat{X}_n) \cdot \xi_{n+1}\sqrt{\Delta t}  - \widehat{V}_{n+1}(\widehat{X}_{n+1})  \big].
\end{align}
\end{subequations}
For $F_1, F_2 \in \widetilde{\mathcal{U}}$ with $F_i = f_i(\widehat{X}_n)$, $f_i \in \mathcal{U}$, we see that
\begin{subequations}
\nonumber
\begin{align}
&\left \Vert \Psi F_1 - \Psi F_2 \right \Vert_{L^2(\mathbb{P})}  
\\
&= \big \Vert 
\Pi_{\widetilde{\mathcal{U}}} \big[ -h(\widehat{X}_n,t_n,f_1(\widehat{X}_n),\sigma^\top \nabla f_1 (\widehat{X}_n)) \Delta t
\\
&+h(\widehat{X}_n,t_n,f_2(\widehat{X}_n),\sigma^\top \nabla f_2 (\widehat{X}_n)) \Delta t
\\
&+ \sqrt{\Delta t}\left(\sigma^\top \nabla f_1(\widehat{X}_n) - \sigma^\top \nabla f_2(\widehat{X}_n) \right)\cdot \xi_{n+1}
\big]
\big \Vert_{L^2(\mathbb{P})}
\\
&\le C_3\left \Vert \Pi_{\widetilde{\mathcal{U}}} \right \Vert_{L^2(\mathbb{P}) \rightarrow L^2(\mathbb{P})} \Bigg( \Delta t \Vert F_1 - F_2 \Vert_{L^2(\mathbb{P})}
\\
&+ \sqrt{\Delta t} \left\Vert 
\left(\sigma^\top \nabla f_1(\widehat{X}_n) - \sigma^\top \nabla f_2(\widehat{X}_n) \right)\cdot \xi_{n+1} \right\Vert_{L^2(\mathbb{P})}
\Bigg)
\end{align}
\end{subequations}
for some constant $C_3$ that does not depend on $\Delta t$, where we have used the triangle inequality, the Lipschitz assumption on $h$,  the boundedness of $\sigma$, and the estimate \eqref{eq:der estimate}. Using the Cauchy-Schwarz inequality, boundedness of $\sigma$ as well as \eqref{eq:der estimate} and  \eqref{eq:moment estimate}, the last term can be estimated as follows,
\begin{align}
\nonumber
\left\Vert 
\left(\sigma^\top \nabla f_1(\widehat{X}_n) - \sigma^\top \nabla f_2(\widehat{X}_n) \right)\cdot \xi_{n+1} \right\Vert_{L^2(\mathbb{P})}
\\
\nonumber
\le \left \Vert \left(\sigma^\top \nabla f_1(\widehat{X}_n) - \sigma^\top \nabla f_2(\widehat{X}_n) \right)^2  \right \Vert^{1/2}_{L^2(\mathbb{P})} \left\Vert \xi_{n+1}^2 \right\Vert^{1/2}_{L^2(\mathbb{P})}
\\
\nonumber
\le C_4 \left \Vert F_1 - F_2\right \Vert_{L^2(\mathbb{P})}, 
\end{align}
where $C_4$ is a constant independent of $\Delta t$. Collecting the previous estimates, we see that $\delta > 0$ can be chosen such that for all $t \in (0,\delta)$, the mapping $\Psi$ is a contraction on $\widetilde{\mathcal{U}}$ when equipped with the norm $\Vert \cdot \Vert_{L^2(\mathbb{P})}$, that is,
\begin{equation}
\Vert \Psi F_1 - \Psi F_2 \Vert \le \lambda \Vert F_1 - F_2 \Vert,
\end{equation}
for some $\lambda <1$ and all $F_1,F_2 \in \widetilde{\mathcal{U}}$. Finally, the statement follows from the Banach fixed point theorem.
\end{proof}

\section{Implementation details}\label{sec:implementation}
\label{app: implementation details}

For the evaluation of our approximations we rely on reference values of $V(x_0, 0)$ and further define the following two loss metrics, which are zero if and only if the PDE is fulfilled along the samples generated by the discrete forward SDE \eqref{eq: discrete SDE}. In the spirit of \cite{raissi2019physics}, we define the \textit{PDE loss} as
\begin{align}
\begin{split}
    &\mathcal{L}_\mathrm{PDE} = \frac{1}{K N}\sum_{n=1}^{N} \sum_{k=1}^K\Big( (\partial_t + L) V(\widehat{X}_n^{(k)}, t_n) \\
    & \qquad + h(\widehat{X}_n^{(k)}, t_n, V(\widehat{X}_n^{(k)}, t_n), (\sigma^\top \nabla V)(\widehat{X}_n^{(k)}, t_n))\Big)^2,
\end{split}
\end{align}
where $\widehat{X}_n^{(k)}$ are realizations of \eqref{eq: discrete SDE}, the time derivative is approximated with finite differences and the space derivatives are computed analytically (or with automatic differentiation tools). 
We leave out the first time step $n=0$ since the regression problem within the explicit and the implicit schemes for the tensor trains are not well-defined due to the fact that $\widehat X_0^k = x_0$ has the same value for all $k$. We still obtain a good approximation since the added regularization term brings a minimum norm solution with the correct point value $V(x_0, 0)$. Still, this does not aim at the PDE being entirely fulfilled at this point in time.

Further, we define the \textit{relative reference loss} as
\begin{align}
    \mathcal{L}_\mathrm{ref} = \frac{1}{K (N+1)}\sum_{n=0}^{N} \sum_{k=1}^K\left| \frac{V(\widehat{X}_n^{(k)}, t_n)  - V_{\text{ref}}(\widehat{X}_n^{(k)}, t_n)}{V_{\text{ref}}(\widehat{X}_n^{(k)}, t_n)} \right|,
\end{align}
whenever a reference solution for all $x$ and $t$ is available.

All computation times in the reported tables are measured in seconds.

Our experiments have been performed on a desktop computer containing an AMD Ryzen Threadripper $2990$ WX $32$x $3.00$ GHz mainboard and an NVIDIA Titan RTX GPU, where we note that only the NN optimizations were run on this GPU, since our TT framework does not include GPU support. It is expected that running the TT approximations on a GPU will improve time performances in the future \cite{abdelfattah2016high}.

All our code is available under \url{https://github.com/lorenzrichter/PDE-backward-solver}.

\subsection{Details on neural network approximation}

For the neural network architecture we rely on the \textit{DenseNet}, which consists of fully-connected layers with additional skip connections as for instance suggested in \cite{weinan2018deepRitz} and being rooted in \cite{huang2017}. To be precise, we define a version of the \textit{DenseNet} that includes the terminal condition of the PDE \eqref{eq: definition general PDE} as an additive extension by
\begin{equation}
\Phi_\varrho(x) = A_L x_L + b_L + \theta g(x),
\end{equation}
where $x_{L}$ is specified recursively as
\begin{align}
y_{l+1} = \varrho(A_l x_l + b_l), \qquad  x_{l+1} = (x_l, y_{l+1})^\top
\end{align}
for $1 \le l \le L-1$ with $A_l \in \R^{r_l \times \sum_{i=0}^{l-1} r_i}, b_l \in \R^l, \theta \in \R$ and $x_1 = x$. The collection of matrices $A_l$, vectors $b_l$ and the coefficient $\theta$ comprises the learnable parameters, and we introduce the vector $r := (d_\text{in}, r_1, \dots, r_{L-1}, d_\text{out})$ to represent a certain choice of a DenseNet architecture, where in our setting $d_\text{in} = d$ and $d_\text{out} = 1$. If not otherwise stated we fix the parameter $\theta$ to be $1$. For the activation function $\varrho: \R \to \R$, that is to be applied componentwise, we choose $\tanh$.

For the gradient descent optimization we choose the Adam optimizer with the default parameters $\beta_1 = 0.9, \beta_2 = 0.999, \varepsilon = 10^{-8}$ \cite{kingma2014adam}. In most of our experiments we chose a fixed learning rate $\eta_{N-1}$ for the approximation of the first backward iteration step to approximate $\widehat{V}_{N-1}$ and another fixed learning rate $\eta_n$ for all the other iteration steps to approximate $\widehat{V}_{n}$ for $0 \le n \le N-2$ (cf. Remark \ref{rem: parameter initializations}). Similarly, we denote with $G_{N-1}$ and $G_n$ the amount of gradient descent steps in the corresponding optimizations.

In Tables \ref{tbl: NN hyperparameters} and \ref{tbl: NN hyperparameters additional experiments} we list our hyperparameter choices for the neural network experiments that we have conducted.

\begin{table}[h!]

\centering
\begin{tabular}{ c }
 HJB, $d = 10$, $\text{NN}_\text{impl}$ \\
Figure \ref{fig: HJB trajectories, d = 10} \\
 \hline
  $K = 2000, \Delta t = 0.01$ \\
 $r = (100, 110, 110, 50, 50, 1)$  \\
 $ G_n = 8000, G_{N-1} = 40000$  \\
 $\eta_{n} = 0.0001, \eta_{N-1} = 0.0001$   \\
 \vspace{0.1cm}
\end{tabular}

\begin{tabular}{ c }
 HJB, $d = 100$, $\text{NN}_\text{impl}$ \\
Table \ref{tbl: HJB d = 100}, Figures \ref{fig: HJB trajectories}, \ref{fig: HJB mean relative error} \\
 \hline
  $K = 2000, \Delta t = 0.01$ \\
 $r = (100, 130, 130, 70, 70, 1)$  \\
 $ G_n = 5000, G_{N-1} = 40000$  \\
 $\eta_{n} = 0.0001, \eta_{N-1} = 0.0003$   \\
 \vspace{0.1cm}
\end{tabular}

\centering
\begin{tabular}{ c }
 HJB, $d = 100$, $\text{NN}_\text{expl}$ \\
Table \ref{tbl: HJB d = 100}, Figures \ref{fig: HJB trajectories}, \ref{fig: HJB mean relative error} \\
 \hline
  $K = 2000, \Delta t = 0.01$ \\
 $r = (100, 110, 110, 50, 50, 1)$  \\
 $ G_n = 500, G_{N-1} = 7000$  \\
 $\eta_{n} = 0.00005, \eta_{N-1} = 0.0003$   \\
 \vspace{0.1cm}
\end{tabular}

\centering
\begin{tabular}{ c }
 HJB double well \\
$d = 50$, $\text{NN}_\text{impl}$, Table \ref{tbl: HJB double well diagonal} \\
 \hline
  $K = 2000, \Delta t = 0.01$ \\
 $r = (50, 30, 30, 1)$  \\
 $ G_n = 2000, G_{N-1} = 25000$  \\
 $\eta_{n} = 0.0002, \eta_{N-1} = 0.0005$   \\
 \vspace{0.1cm}
\end{tabular}

\centering
\begin{tabular}{ c }
 HJB interacting double well \\
$d = 20$, $\text{NN}_\text{impl}$, Table \ref{tbl: HJB double well nondiagonal} \\
 \hline
  $K = 2000, \Delta t = 0.01$ \\
 $r = (50, 20, 20, 20, 20, 1)$  \\
 $ G_n = 3000, G_{N-1} = 30000$  \\
 $\eta_{n} = 0.0007, \eta_{N-1} = 0.001$   \\
 \vspace{0.1cm}
\end{tabular}

\centering
\begin{tabular}{ c }
 CIR, $d = 100$, $\text{NN}_\text{impl}$ \\ 
Table \ref{tbl: cir results} \\
 \hline
 $K = 1000, \Delta t = 0.01$ \\
 $r = (100, 110, 110, 50, 50, 1)$  \\
 $ G_n = 2000$ for $0 \le n \le 15$\\
 $ G_n = 300$ for $16 \le n \le N-2$\\
 $ G_{N-1} = 10000$\\
 $\eta_{n} = 0.00005, \eta_{N-1} = 0.0001$   \\
  \vspace{0.1cm}
\end{tabular}

\caption{Neural network hyperparameters for the experiments in paper.}
\label{tbl: NN hyperparameters}
\end{table}

\begin{table}[h!]
\centering
\begin{tabular}{ c }
 PDE with unbounded solution \\ 
$d = 10$, $\text{NN}_\text{impl}$, Table \ref{tbl: unbounded sin results, d = 10} \\
 \hline
 $K = 1000, \Delta t = 0.001$ \\
 $r = (10, 30, 30, 1)$  \\
 $ G_n = 100, G_{N-1} = 10000$\\
 $\eta_{n} = 0.0001, \eta_{N-1} = 0.0001$   \\
   \vspace{0.1cm}
\end{tabular}

\centering
\begin{tabular}{ c }
 Allen-Cahn \\ 
$d = 100$, $\text{NN}_\text{impl}$, Table \ref{tbl: Allen-Cahn results delta t = 0.01} \\
 \hline
 $K = 8000, \Delta t = 0.01$ \\
 $r = (10, 30, 30, 1)$  \\
 $ G_n = 10000$ for $0 \le n \le 5$\\
 $ G_n = 6000$ for $6 \le n \le N-2$\\
 $ G_{N-1} = 15000$\\
 $\eta_{n} = 0.0002, \eta_{N-1} = 0.001$   \\
\end{tabular}
\caption{Neural network hyperparameters for the additional experiments.}
\label{tbl: NN hyperparameters additional experiments}
\end{table}

\subsection{Details on tensor train approximation}\label{app:tt_details}
For the implementation of the tensor networks we rely on the C++ library \emph{xerus} \cite{xerus} and the Python library \emph{numpy} \cite{harris2020array}.

Within the optimization we have to specify the regularization parameter as noted in Remark \ref{rem:regularization}, which we denot here by $\eta > 0$.
We adapt this parameter in dependence of the current residual in the regression problem \eqref{eq:fixed-point}, i.e. $\eta = c w$, where $c > 0$ and $w$ is the residual from the previous sweep of SALSA. 
In every all our experiments we set $c_\eta = 1$.
Further, we have to specify the condition ``\emph{noChange} is \emph{true}'' within Algorithm \ref{alg:salsa}.
To this end we introduce a test set with equal size as our training set.
We measure the residual within a single run of SALSA on the test set and the training set.
If the change of the residual on either of this sets is below $\delta = 0.0001$ we set \emph{noChange = true}.
For the fixed-point iteration we have a two-fold stopping condition.
We stop the iteration if either the Frobenius norm of the coefficients has a smaller relative difference than $\gamma_1 < 0.0001$ or if the values $\widehat V_n^{k+1}$ and $\widehat V_n^k$ and their gradients, evaluated at the points of the test set, have a relative difference smaller than $\gamma_2 < 0.00001$.
Note that the second condition is essentially a discrete $H^1$ norm, which is necessary since by adding the final condition into the ansatz space the orthonormal basis property is violated.

\review{Finally, we comment on the area $[a, b]$ where the 1-dimensional polynomials are orthonormalized w.r.t. the $H^2(a, b)$ norm, c.f. Remark \ref{rem:regularization}.
We obtain these polynomials by performing a Gram-Schmidt process starting with one-dimensional monomials.
Thus, we have to specify the integration area $[a, b]$ for the different tests.
In Section \ref{sec: HJB equation} we set $a=-6$ and $b = 6$. In Section \ref{sec: HJB double well} we set $a = -3$ and $b = 3$ for the case $C$ diagonal and for the interacting case, where $C$ is non-diagonal, we set $a = -8$ and $b = 2$.
In Section \ref{sec:cir} we choose $a = -0.2$ and $b = 6$.}

\section{Further numerical examples}
\label{app: further numerical examples and details}

In this section we elaborate on some of the numerical examples from the paper and provide two additional problems.

\subsection{Hamilton-Jacobi-Bellman equation}
\label{app: HJB}

Let us consider the HJB equation from Sections \ref{sec: HJB equation} and \ref{sec: HJB double well}, which we can write as 
\begin{subequations}
\begin{align}
    \left(\partial_t + L \right)V(x, t) - \frac{1}{2}|(\sigma^\top \nabla V)(x, t)|^2 &= 0, \\
    V(x, T) &= g(x),
\end{align}
\end{subequations}
in a generic form with the differential operator $L$ being defined in \eqref{eq: infinitesimal generator}. We can introduce the exponential transformation $\psi := e^{-V}$ and with the chain rule find that the transformed function fulfills the linear PDE
\begin{subequations}
\begin{align}
    \left(\partial_t + L \right)\psi(x, t) &= 0, \\
    \psi(x, T) &= e^{-g(x)}.
\end{align}
\end{subequations}
This is known as Hopf-Cole transformation, see also \cite{fleming2006controlled, hartmann2017variational}. It is known that via the Feynman-Kac theorem \cite{karatzas1998brownian} the solution to this PDE has the stochastic representation
\begin{equation}
    \psi(x, t) = \E\left[e^{-g(X_T)} \Big| X_t = x \right],
\end{equation}
such that we readily get 
\begin{equation}
    V(x, t) = -\log \E\left[e^{-g(X_T)} \Big| X_t = x \right],
\end{equation}
which we can use as a reference solution by approximating the expectation value via Monte Carlo simulation, however keeping in mind that in high dimensions corresponding estimators might have high variances \cite{hartmann2021}.

\begin{figure}[h!]
\vskip 0.2in
\begin{center}
\centerline{\includegraphics[width=1.0\columnwidth]{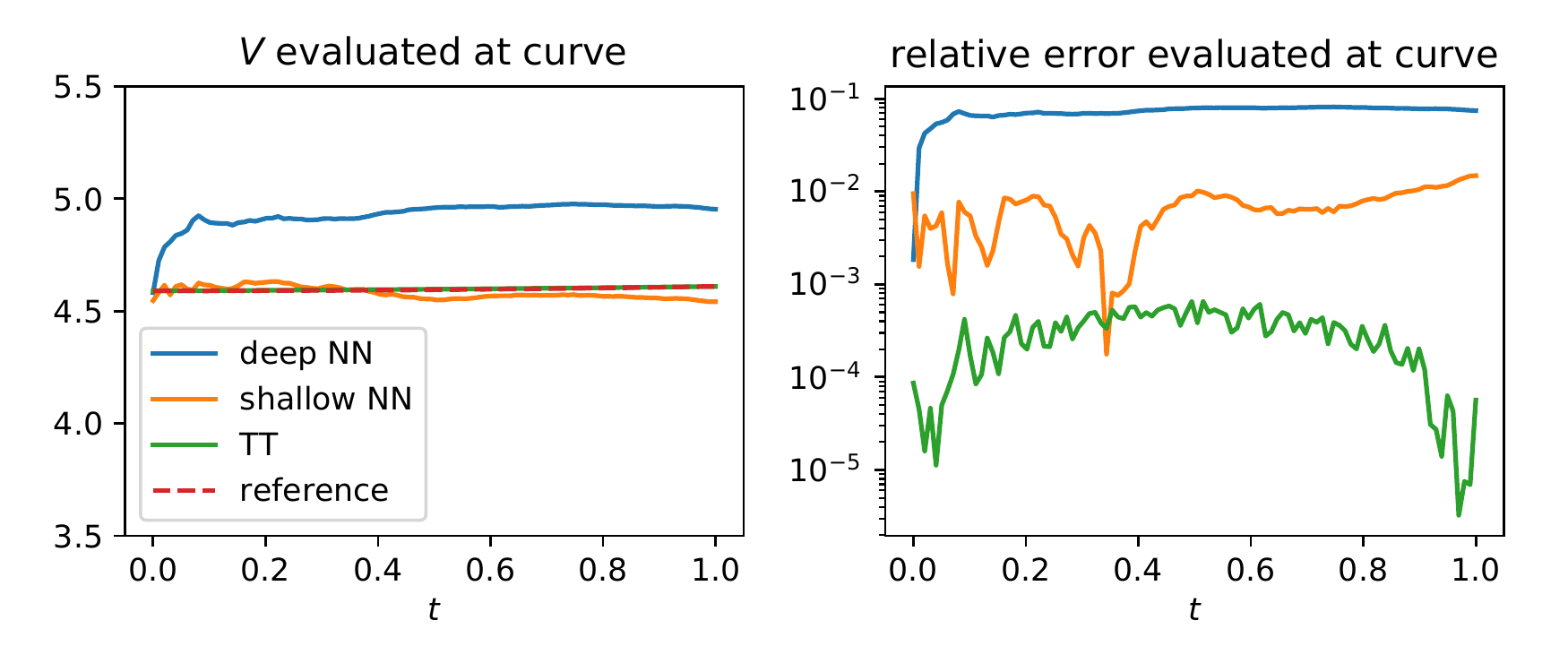}}
\caption{Approximations of the HJB equation in $d=100$ evaluated along a representative curve.}
\label{fig: HJB evaluation along curve}
\end{center}
\vskip -0.2in
\end{figure}

Let us stress again that our algorithms only aim to provide a solution of the PDE along the trajectories of the forward process \eqref{eq: fordward SDE}. Still, there is hope that our approximations generalize to regions ``close'' to where samples are available. To illustrate this, consider for instance the $d$-dimensional forward process
\begin{equation}
    X_s = x_0 + \sigma W_s,
\end{equation}
as for instance in Section \ref{sec: HJB equation}, where now $\sigma > 0$ is one-dimensional for notational convenience. We know that $X_t \sim \mathcal{N}(x_0, \sigma^2 t \,\mathrm{Id}_{d \times d})$ and therefore note that for the expected distance to the origin it holds
\begin{align}
    \E\left[|X_t - x_0| \right] < \sqrt{\E\left[|X_t - x_0|^2 \right]} = \sigma \sqrt{d  t}.
\end{align}
This motivates evaluating the approximations along the curve
\begin{equation}
    X_t = x_0 + \sigma \sqrt{t}\mathbf{1},
\end{equation}
where $\mathbf{1} = (1, \dots, 1)^\top$. Figure \ref{fig: HJB evaluation along curve} shows that in this case we indeed have good agreement of the approximation with the reference solution when using TTs and that for NNs the deep neural network that we have specified in Table \ref{tbl: NN hyperparameters} generalizes worse than a shallower network with only two hidden layers consisting of $30$ neurons each.

\subsection{PDE with unbounded solution}

As an additional problem, we choose an example from \cite{hure2020deep} which offers an analytical reference solution. For the PDE as defined in \eqref{eq: definition general PDE} we consider the coefficients
\begin{gather}
b(x, t) = \mathbf{0}, \,\,\sigma(x, t) = \frac{\mathrm{Id}_{d \times d}}{\sqrt{d}} ,\,\, g(x) = \cos \left(\sum_{i=1}^d i x_i \right),\\
    h(x, t, y, z) =k(x) + \frac{y}{2 \sqrt{d}} \sum_{i=1}^d z_i + \frac{y^2}{2},
\end{gather}
where, with an appropriately chosen $k$, a solution can shown to be
\begin{align}
\begin{split}
    V(x, t) &= \frac{T - t}{d} \sum_{i=1}^d\left(\sin(x_i) \mathds{1}_{x_i < 0} + x_i \mathds{1}_{x_i \ge 0} \right) \\
    &\qquad + \cos \left(\sum_{i=1}^d i x_i \right).
\end{split}
\end{align}

In Table \ref{tbl: unbounded sin results, d = 10} we compare the results for $d=10, K=1000, T = 1, \Delta t = 0.001, x_0 = (0.5, \dots, 0.5)^\top$.
For the TT case it was sufficient to set the ranks to $1$ and \review{the polynomial degree to $6$.} We see that the results are improved significantly if we increase the sample size $K$ from $1000$ to $20000$.
Note that even when increasing the sample size by a factor $20$, the computational time is still lower than the NN implementation.
It should be highlighted that adding the function $g$ to the neural network (as explained in Appendix \ref{app: implementation details}) is essential for its convergence in higher dimensions and thereby mitigates the observed difficulties in \cite{hure2020deep}).

\begin{table}[h!]
\centering
\begin{tabular}{ c | c  c  c }
& $\text{TT}_\text{impl}$ & $\text{TT}_\text{impl}^*$ & $\text{NN}_\text{impl}$ \\
 \hline
$\widehat{V}_0(x_0)$ & $-0.1887$ & $-0.2136$  &  $-0.2137$ \\
relative error & $1.22\text{e}^{-1}$ & $6.11\text{e}^{-3}$  & $5.50\text{e}^{-3}$ \\
ref loss & $2.47\text{e}^{-1}$ & $7.57\text{e}^{-2}$ &  $3.05\text{e}^{-1}$ \\
abs. ref loss & $2.52\text{e}^{-2}$ & $9.29\text{e}^{-3}$  &  $1.69\text{e}^{-2}$ \\
PDE loss & $2.42$ & $0.60$  & $1.38$\\
computation time & $360$ &  $1778$  & $4520$ \\
\end{tabular}
\caption{Approximation results for the PDE with an unbounded analytic solution. For $\text{TT}_\text{impl}^*$ we choose $K=20000$, for the others we choose $K=1000$.}
\label{tbl: unbounded sin results, d = 10}
\end{table}

\subsection{Allen-Cahn like equation}

Finally, let us consider the following Allen-Cahn like PDE with a cubic nonlinearity in $d=100$:
\begin{subequations}
\begin{align}
    (\partial_t + \Delta) V(x, t) + V(x, t) - V^3(x, t) &= 0, \\
    V(x, T) &= g(x),
\end{align}
\end{subequations}
where we choose $g(x) = \left(2 + \frac{2}{5}|x|^2\right)^{-1}$, $T = \frac{3}{10}$ and are interested in an evaluation at $x_0 = (0, \dots, 0)^\top$. This problem has been considered in \cite{weinan2017deep}, where a reference solution of $V(x_0, 0) = 0.052802$ calculated
by means of the branching diffusion method is provided. We consider a sample size of $K = 1000$ and a stepsize $\Delta t = 0.01$ and provide our approximation results in Table \ref{tbl: Allen-Cahn results delta t = 0.01}. \review{Note that for this example it is again sufficient to use a TT-rank of $1$ and a polynomial degree of $0$.}

\begin{table}[h!]
\centering
\begin{tabular}{ c | c  c  c  c }
& $\text{TT}_\text{impl}$ & $\text{TT}_\text{expl}$ & $\text{NN}_\text{impl}$ & $\text{NN}^*_\text{impl}$ \\
 \hline
$\widehat{V}_0(x_0)$ & $0.052800$ & $0.05256$ & $0.04678$ & $0.05176$ \\
relative error & $4.75\text{e}^{-5}$ & $4.65\text{e}^{-3}$ & $1.14\text{e}^{-1}$ & $1.97\text{e}^{-2}$ \\
PDE loss & $2.40 \text{e}^{-4}$ & $2.57\text{e}^{-4}$ & $9.08\text{e}^{-1}$ & $6.92\text{e}^{-1}$ \\
comp. time & $24$ & $10$ & $23010$& $95278$ \\
\end{tabular}
\caption{Approximations for Allen-Cahn PDE, where $\text{NN}^*_\text{impl}$ uses $K=8000$ and the others $K =1000$ samples.}
\label{tbl: Allen-Cahn results delta t = 0.01}
\end{table}

\section{Some background on BSDEs and their numerical discretizations}
\label{app: BSDE background}

BSDEs have been studied extensively in the last three decades and we refer to \cite{pardoux1998backward, pham2009continuous, gobet2016monte, zhang2017backward} for good introductions to the topic. Let us note that given some assumptions on the coefficients $b, \sigma, h$ and $g$ one can prove existence and uniqueness of a solution to the BSDE system as defined in \eqref{eq: fordward SDE} and \eqref{eq: BSDE}, see for instance Theorem 4.3.1 in \cite{zhang2017backward}.

We note that the standard BSDE system can be generalized to
\begin{subequations}
\begin{align}
    \mathrm dX_s &= \left(b(X_s, s) + v(X_s, s)\right) \mathrm ds + \sigma(X_s, s) \mathrm d W_s,\\
     X_0 &= x, \\
     \mathrm dY_s &= (-h(X_s, s, Y_s, Z_s)  + v(X_s, s)\cdot Z_s) \mathrm ds + Z_s \cdot \mathrm d W_s,\\
     Y_T &= g(X_T),
\end{align}
\end{subequations}
where $v : \R^d \times [0, T] \to \R^d$ is any suitable control vector field that can be understood as pushing the forward trajectories into desired regions of the state space, noting that the relations 
\begin{equation}
    Y_s = V(X_s, s), \qquad Z_s = (\sigma^\top \nabla V)(X_s, s),
\end{equation}
with $V : \R^d \times [0, T] \to \R$ being the solution to the parabolic PDE \eqref{eq: definition general PDE}, hold true independent of the choice of $v$ \cite{hartmann2019variational}. Our algorithms readily transfer to this change in sampling the forward process by adapting the backward process and the corresponding loss functionals \eqref{eq: projection_based_optimization} and \eqref{eq: implicit scheme} accordingly. 

In order to understand the different numerical discretization schemes in Section \ref{sec: numerical approximation of BSDEs}, let us note that we can write the backward process \eqref{eq: def Y Z} in its integrated form for the times $t_n < t_{n+1}$ as
\begin{equation}
    Y_{t_{n+1}} = Y_{t_n} - \int_{t_n}^{t_{n+1}} h(X_s, s, Y_s, Z_s) \mathrm ds + \int_{t_n}^{t_{n+1}} Z_s \cdot \mathrm d W_s.
\end{equation}
In a discrete version we have to replace the integrals with suitable discretizations, where for the deterministic integral we can decide which endpoint to consider, leading to either of the following two discretization schemes
\begin{subequations}
\begin{align}
    \label{eq: discrete BSDE appendix}
    \widehat{Y}_{n+1} &= \widehat{Y}_n - h_n \Delta t +  \widehat{Z}_n \cdot \xi_{n+1}  \sqrt{\Delta t}, \\
    \label{eq: discrete BSDE explicit appendix}
    \widehat{Y}_{n+1} &= \widehat{Y}_n - h_{n+1} \Delta t +  \widehat{Z}_n \cdot \xi_{n+1}  \sqrt{\Delta t},
\end{align}
\end{subequations}
as defined in \eqref{eq: backward schemes}, where we recall the shorthands
\begin{subequations}
\begin{align}
    h_n &= h(\widehat{X}_n, t_n, \widehat{Y}_n, \widehat{Z}_n), \\ 
    h_{n+1} &= h(\widehat{X}_{n+1}, t_{n+1}, \widehat{Y}_{n+1}, \widehat{Z}_{n+1}).
\end{align}
\end{subequations}

The $L^2$-projection scheme \eqref{eq: projection_based_optimization} can be motivated as follows. Consider the explicit discrete backward scheme as in \eqref{eq: discrete BSDE explicit appendix}
\begin{equation}
        \widehat{Y}_{n+1} = \widehat{Y}_n - h(\widehat{X}_{n+1}, t_{n+1}, \widehat{Y}_{n+1}, \widehat{Z}_{n+1})\Delta t +  \widehat{Z}_n \cdot \xi_{n+1}  \sqrt{\Delta t}.
\end{equation}
Taking conditional expectations w.r.t. to the $\sigma$-algebra generated by the discrete Brownian motion at time step $n$, denoted by $ \mathcal{F}_n$, yields
\begin{equation}
        \widehat{Y}_{n} = \E\left[\widehat{Y}_{n+1} + h(\widehat{X}_{n+1}, t_{n+1}, \widehat{Y}_{n+1}, \widehat{Z}_{n+1})\Delta t \Big| \mathcal{F}_n \right].
\end{equation}
We can now recall that a conditional expectation can be characterized as a best approximation in $L^2$, namely
\begin{equation}
\E[B  | \mathcal{F}_n] = \argmin_{\substack{Y \in L^2 \\ \mathcal{F}_n-\text{measurable}}}\E\left[|Y-B|^2\right],
\end{equation}
for any random variable $B \in L^2$, which brings
\begin{equation}
\widehat{Y}_{n} = \argmin_{\substack{Y \in L^2 \\ \mathcal{F}_n-\text{measurable}}} \E\left[ \left(Y  - h_{n+1} \Delta t - \widehat{Y}_{n+1} \right)^2\right].
\end{equation}
This then yields the explicit scheme depicted in \eqref{eq: projection_based_optimization}. We refer once more to \cite{gobet2005regression} for extensive numerical analysis, essentially showing that the proposed scheme is of order $\frac{1}{2}$ in the time step $\Delta t$.

\end{document}